\documentclass[twocolumn]{article}[12]

\usepackage{PRIMEarxiv}
\usepackage[utf8]{inputenc} 
\usepackage[T1]{fontenc}    
\usepackage{hyperref}       
\usepackage{url}            
\usepackage{booktabs}       
\usepackage{amsfonts}       
\usepackage{nicefrac}       
\usepackage{microtype}      
\usepackage{lipsum}
\usepackage{fancyhdr}       
\usepackage{graphicx}       
\usepackage{amsmath}
\usepackage{dsfont}
\usepackage{bm} 

\usepackage{amsfonts}
\usepackage{amssymb}
\usepackage{caption}
\usepackage{subcaption}
\usepackage{booktabs}
\usepackage{multirow}
\usepackage{graphicx}
\usepackage{xcolor}
\usepackage{svg}
\usepackage[thmmarks, amsmath, amsthm]{ntheorem}

\DeclareGraphicsExtensions{.png,.pdf}

\newcommand*{\vertbar}{\rule[-1ex]{0.5pt}{2.5ex}}
\newcommand*{\horzbar}{\rule[.5ex]{2.5ex}{0.5pt}}

\DeclareMathOperator*{\argmin}{arg\,min}
\newcommand*{\Resize}[2]{\resizebox{#1}{!}{$#2$}}%

\usepackage{algorithmic}
\usepackage{algorithm}

\title{Biomechanical Constraints Assimilation in Deep-Learning Image Registration: Application to sliding and locally rigid deformations}

\author{
Ziad Kheil$^{1,2,3}$, Soleakhena Ken $^{1,3}$, Laurent Risser$^{2}$\\
$\null$\\
  $\null^{1}$ Centre de Recherches en Canc\'erologie de Toulouse, INSERM UMR1037 \\
$\null^{2}$ Institut de Math\'ematiques de Toulouse (UMR 5219), CNRS, Universit\'e de Toulouse, F-31062 Toulouse, France\\
  $\null^{3}$ Institut Universitaire du Cancer – Oncopole Claudius R\'egaud, 31059 Toulouse, France \\
}

\begin{document}

\twocolumn[
  \begin{@twocolumnfalse}
    \maketitle
    \begin{abstract}

Regularization strategies in medical image registration often take a one-size-fits-all approach by imposing uniform constraints across the entire image domain. Yet biological structures are anything but regular.
Lacking structural awareness, these strategies may fail to consider a panoply of spatially inhomogeneous deformation properties, which would faithfully account for the biomechanics of soft and hard tissues, especially in poorly contrasted structures.

To bridge this gap, we propose a learning-based image registration approach in which the inferred deformation properties can locally adapt themselves to trained biomechanical characteristics. Specifically, we first enforce in the training process local rigid displacements, shearing motions or pseudo-elastic deformations using regularization losses inspired from the field of solid-mechanics. We then show  on synthetic and real 3D thoracic and abdominal images that these mechanical properties of different nature are well generalized when inferring the deformations between new image pairs. Our approach enables neural-networks to infer tissue-specific deformation patterns directly from input images, ensuring mechanically plausible motion. These networks preserve rigidity within hard tissues while allowing controlled sliding in regions where tissues naturally separate, more faithfully capturing physiological motion. The code is publicly available at \url{https://github.com/Kheil-Z/biomechanical_DLIR}.
\vspace{1cm}
\end{abstract}
  \end{@twocolumnfalse}
]

\keywords{medical image registration; neural-networks; biomechanical  constraints; 3D thoracic images}

\section{Introduction}\label{sec:intro}

Deformable Image Registration (DIR) plays an essential role in most modern medical imaging applications. DIR attempts to align images by allowing flexible, non-linear transformations to accurately map anatomical or structural variations. 
As an inverse problem, it is often solved by regularized optimization for a pair of given images, which is known as the \textit{iterative formulation}. More recently, the iterative formulation of DIR has been outpaced by the introduction of Deep Learning Image Registration (DLIR), which leverages the expressive power of Deep Neural Networks (DNN) to predict adequate spatial correspondences given the images to register with short inference times. Although training neural networks for accurate DLIR can be demanding in terms of computational resources, its predictions can indeed be made in seconds on conventional computers. 
Both DIR and DLIR can now be considered as mature problems, as substantial research has been done on regularization strategies, deformation parametrization, optimization strategies, neural architecture and use cases \cite{sotiras_deformable_2013,boveiri_medical_2020,jena_deep_2024,HEINRICH2020293}. However, our work deals with spatial regularization in DLIR, for which we believe that important open questions remain to properly take into account local biomechanical properties of deformed anatomical structures. Unlike classic iterative solutions, where spatial smoothness is obtained either by design or by using hard-coded regularization models, DLIR implicitly makes use of knowledge trained on paired image sets to spatially regularize the predicted deformations between two registered images.
An extreme illustration of this claim is \cite{HoffmannTMI2022}, where the authors trained DNNs to perform multimodal brain image registration with synthetic images that do not look like brains but present locally realistic contrast and geometrical features.

\begin{figure}[thb]
\centering
\includegraphics[width=0.99\columnwidth]{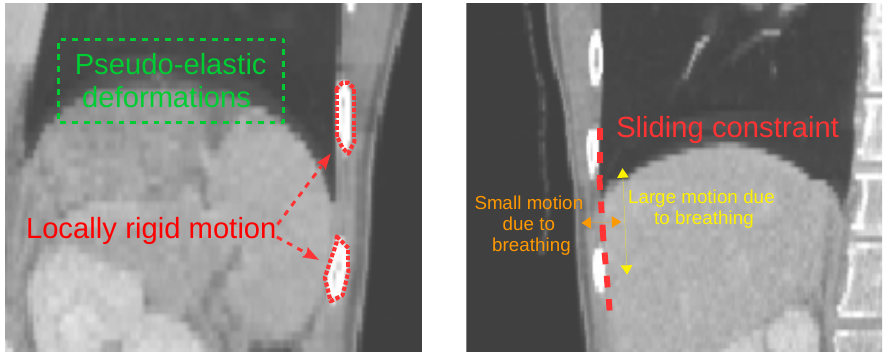}
\caption{Possible hypotheses on the movement of different anatomical structures in medical image registration. Pseudo-elastic deformations are classically used throughout the image domain. However, it may be advantageous to consider locally more realistic biomechanical constraints on the deformed structures. \textbf{(left)} Locally rigid deformations can reasonably be considered in bones, as here in the ribs. \textbf{(right)} Sliding motion may also occur, as here at the thoracic cage boundary, next to the lungs/liver interface. Breathing indeed induces there a large up and down motion in the inner part of the thoracic cage, while the ribs are not impacted by this motion. Importantly, thoracic cage swelling also induces a small displacement of the locations where sliding phenomena occur. This motion is globally orthogonal to the large up and down motion.}\label{fig:motivating_example}
\end{figure}

In other supervised learning strategies, the observations on which the predictions are made must follow a statistical distribution similar to the data on which the prediction rules are trained. In the medical DLIR context, this means that the true anatomical deformations, that are expected to be found by the DLIR algorithm, should be similar to the motion captured in the training images.
As illustrated Figure~\ref{fig:motivating_example}, this motion may be biomechanically constrained by deformation models that are spatially inhomogeneous and have a different nature (\textit{e.g.} elasticity, rigidity, sliding motion). 
Two central questions addressed in this paper are then: (1) how can we help a DLIR algorithm to learn bio-mechanically realistic deformations in the training images? (2) Will a DLIR algorithm be able to generalize such realistic deformations?  These questions appear to us as critical in a context where the size of the open training bases \cite{Zhoubing_learn2reg} or internal clinical dataset for training is often much lower than the amount of degrees of freedom to model the image mappings. Being able to inject and generalize a knowledge on the bio-mechanical properties of the deformed structures is therefore a natural approach to address the curse-of-dimensionality problem there.

Specifically, the main methodological contribution of this paper is the design of novel structure dependent DLIR losses for locally rigid deformations (\textit{e.g.} as in the bones), sliding motion (\textit{e.g.} as at the interface between the lungs and the liver or the ribs) or default pseudo-elasticity. The main interest of these losses is that they must be used to bio-mechanically inform the registration during training, but no specific procedure has to be followed when routinely using the trained DLIR system. We also show how to simultaneously use these losses of different nature at different spatial locations.
We will apply our methods to thoraco-abdominal registration, where certain anatomically plausible deformations can hardly be addressed by current DIR or DLIR strategies. 

We first present related works from several perspectives in Section~\ref{sec:related_work}. We then give the interpretation of different solid-mechanics issued modeling strategies in Section~\ref{sec:rigid_mec_tools} and incorporate them to DNN losses in Section~\ref{sec:Mec_losses}. This leads to a practical strategy, which is presented in Section~\ref{sec:annot_images} to automatically parametrize these losses on 3D CT thoraco-abdominal images. Results are finally shown and discussed in Section~\ref{sec:materials_methods}.

\section{Related Work}
\label{sec:related_work}

Across subsections~\ref{sec:deformable_image_regsitration} and \ref{sec:mecanical_constraints}, we respectively give a concise overview of DLIR and of the incorporation of biomechanical constraints in DLIR. For completeness, we also give in Subsection \ref{ssec:PINNS_others} a brief overview of important contributions dealing with the incorporation of mechanics-informed constraints in other applications of DNNs.


Readers seeking a detailed exploration of medical image registration can refer to \cite{maintz_survey_1998, sotiras_deformable_2013, oliveira_medical_2014}, which cover the topic comprehensively from the scope of iterative methods, meanwhile \cite{fu_deep_2020,bharati_deep_2022, chen_survey_2025} detail modern deep neural network approaches. 
We also emphasize that we focus on the incorporation of biomechanical constraints in DNN-based DIR, because these strategies have much faster inference times than when registering 3D medical images with traditional DIR algorithms. A wide literature however exists on the incorporation of bio-mechanical constraints in traditional DIR \cite{WernerMedPhy2009,SCHMIDTRICHBERG2012150,RISSER2013182,PAPIEZ20141299,Lacroix_2021_CVPR}.

\subsection{Deformable image registration with DNNs}
\label{sec:deformable_image_regsitration}

\subsubsection{General notations for DIR}

Given two 3D images $I_M$ and $I_F$ defined over $\Omega_F, \Omega_M \subset \mathbb{R}^3$, DIR consists in obtaining a transformation $\Phi$ that projects the coordinates of the \textit{moving} image $I_M$ to their \textit{correct} positions on the \textit{fixed} image $I_F$ through a displacement field $\Phi$. 
The produced image, known as the \textit{deformed} image can be obtained by composition of the mapping at every coordinate of the fixed image: $I_M \circ \Phi$ which aligns with $I_F$.

The alignment quality between $I_M \circ \Phi$ and $I_F$ is measured using an similarity metric $\mathcal{L}_{sim}$, which gives a function to be minimized when solving the DIR problem. Commonly used similarity metrics are mean squared error (MSE), normalized cross-correlation (NCC), and mutual information (MI). Given the ill-posed nature of the problem \cite{fischer_ill-posed_2008}, regularization strategies are also used to make the problem well-posed. Regularization is generally introduced as a penalty measure depending on the deformation field $\mathcal{L}_{reg}(\Phi)$, some of which are further discussed in Section~\ref{sec:mecanical_constraints}. Hence the DIR problem can be written as  
\begin{equation}\label{eq:dir_variational}
    \Phi^{*} = \argmin \limits_{\Phi} \mathcal{L}_{sim}(I_F,I_M \circ \Phi) +\lambda \mathcal{L}_{reg}(\Phi) \,.
\end{equation}

\subsubsection{Deep Learning Image Registration (DLIR)}

In the classical iterative approach, Eq.~\eqref{eq:dir_variational} can be solved by direct optimization for a given pair of images, but recently unsupervised learning approaches have been applied to the problem to speed up registration times and make use of additional available information by leveraging dataset statistical properties. A backbone DNN $g_{\theta}$ is made to produce a mapping $u:\mathbb{R}^n \to \mathbb{R}^n$ given a pair of images $(\Tilde{I_F},\Tilde{I_M})$ from a dataset. Using a weightless Spatial Transformer Network (STN) layer \cite{jaderberg_spatial_2016}, the mapping can be applied to obtain $\Tilde{I_M} \circ \Phi$ in a differentiable manner. Hence, the network can be trained end-to-end in an unsupervised fashion. Given a training dataset $\mathcal{D}_{\textit{train}} = \{ (I_F^i,I_M^i)\}_{i=1}^{n}$ containing $n$ paired images,  the DLIR training objective resembles Eq.~\eqref{eq:dir_variational}, with the key difference that instead of optimizing a parametrized transformation $\Phi$, we optimize the weights $\theta$ of a DNN which produces transformations (\emph{training procedure}):

\begin{equation}
    \theta^{\ast}  = \mathbb{E}_{ \mathcal{D}_{\textit{train}}} \left( \mathcal{L}_{sim}(I_F,I_M \circ g_{\theta}(I_F,I_M)) +\lambda \mathcal{L}_{reg}(g_{\theta}(I_F,I_M)) \right) \,,
    \label{eq:dir_dnn}
\end{equation}
where $\theta$ are the learnable weights of the model $g_{\theta}$, which produces the DDF: $g_{\theta}(I_F,I_M) = \Phi_{I_M \to I_F}$.
After the lengthy training step, two previously unseen images $(I_F^{\prime},I_M^{\prime}) \notin \mathcal{D}_{\textit{train}}$ can be registered immediately by simply computing the forward pass of the network  (\emph{inference procedure}):
\begin{equation}
\Phi_{M^{\prime} \to F^{\prime}} = g_{\theta^{\ast}} \left(I_F^{\prime},I_M^{\prime} \right)  \,.
\end{equation}

The unsupervised training objective in Eq~\eqref{eq:dir_dnn} using STNs was first introduced as VoxelMorph \cite{balakrishnan_unsupervised_2018} using a UNet \cite{navab_u-net_2015} backbone. Building on this foundation, later research has refined and expanded the formulation by introducing diverse backbone DNN architectures \cite{jia_u-net_2022,mok_large_2020,hering_mlvirnet_2019,liu_vector_2024}, loss functions \cite{niethammer_metric_2019,hoopes2021hypermorph}, and optimization strategies \cite{heo_abdominopelvic_2022,xu_deepatlas_2019}. The unsupervised DLIR formulation notably allowed for a great improvement in inference times over iterative optimization schemes, without a loss of performance. Furthermore, \cite{jena_deep_2024} indicates that semi-supervised DLIR outperforms iterative optimization specifically by introducing additional loss terms when additional labels are available for the dataset.  Using a  dice loss when segmentation masks are securable, or target registration errors when matching keypoints are given  allows the networks to generalize desirable properties beyond simple voxel intensity matching objectives.

\subsection{Mechanical smoothness constraints}
\label{sec:mecanical_constraints}

Although semi-supervised DLIR and label-guided traditional DIR introduce  structural awareness to the optimization process, they are insufficient for producing physically acceptable solutions. Explicitly describing realistic and plausible solutions has been a long standing problem in the field of image mapping. Numerous approaches have been proposed, and while we briefly outline the topology of these methods here, we refer interested readers to \cite{reithmeir_model_2024} for a more comprehensive view of regularization in medical image registration.

\subsubsection{Soft Constraints in DIR: Regularization Penalties}
In light of anatomical ambiguities during the optimization process, there is a need to measure and penalize unrealistic/unwanted movements. In classical DIR (and DL approaches likewise) this takes an \textit{explicit form}, i.e an additional regularization term in the objective function $\mathcal{L}_{reg}$ (as introduced in Eq.~\eqref{eq:dir_variational} and Eq.~\eqref{eq:dir_dnn}). Note that throughout the following, we intentionally omit the use of additional loss terms through semi-superivsed DLIR, such as Dice losses to measure structural alignment or keypoint target registration error, although it can be argued that these contributed to smoothing out undesirable solutions.

DIR approaches overwhelmingly rely on a loss function that measures the spatial gradient of the deformation field \cite{reithmeir_model_2024}. More specifically in the medical context, given that tissues deform smoothly, adjacent image locations are required to vary gradually as well. Hence the spatial gradient of $u$ is often penalized using an $L_1$ or $L_2$-norm. Second order gradient penalization terms such as bending energy are also often introduced. Finally, both classical and modern approaches have employed regularization terms which directly measure topology violations such as foldings. This is often obtained through the Jacobian determinant of the deformation \cite{hutchison_diffeomorphic_2006,christensen_consistent_2001}.
In addition to gradient based penalties, several approaches introduce additional explicit regularization terms on the resulting deformation by measuring inverse-consistency or cycle-consistency \cite{Chen2010,greer2021icon,greer2023inverse,estienne2021mics}. That is, a pair of images ($I_1$,$I_2$) are registered in both directions resulting in two deformation fields $\Phi_{1\to2}$ and $\Phi_{2\to1}$. Inverse-consistency is measured as deviations of the composition $\Phi_{1\to2} \circ \Phi_{1\to2}$ and $\Phi_{2\to1} \circ \Phi_{1\to2}$ to the identity deformation. 

While spatial gradient based regularization helps produce physically acceptable solutions, the same property is enforced through the whole deformation field. This introduces additional hyper-parameters on the loss functions which need to be carefully tuned. Certain approaches such as \cite{hoopes2021hypermorph} introduce architectures which produce regularization-dependent deformation fields. Beyond the challenge of tuning this parameter, a fixed regularization loss is still uniformly applied across the entire deformation field, disregarding local variations. Given the diversity of tissues and their properties present in an image, applying a unified regularization level (loss weight $\lambda$ in Eq.~\eqref{eq:dir_dnn} ) suits only certain tissues and over- or under-penalizes certain displacements. 

Spatially varying regularization functions have been proposed to surmount the problems caused by stationary functions.
Several works introduced spatially varying regularizers which can be optimized using conventional methods \cite{kabus_variational,risser_spatially}. In the learning based framework, one way to achieve spatially varying regularization is to extend the scalar regularization weight $\lambda$ in Eq. 1 to a location dependent weight map $\lambda(x) \in \mathcal{R}^{HxWxD}$ \cite{STEFANESCU2004325}, or $\vec{\lambda} = [\lambda_{0},\lambda_{0},\lambda_{1}, \dots ,\lambda_{K-1}]$ for different regions \cite{Wang2023ConditionalDI}. More recently, spatially-adaptive regularizers, which adapt or learn the spatial regularization levels have also been proposed \cite{Wei2021RecurrentTN}. They adapt pioneer works on Adaptive smoothing \cite{cahill_demons_adaptive} to the DLIR framework by learning an unsupervised voxel-wise regularization classification, which assigns each voxel to one of three smoothing levels (strong, medium, weak). Unlike traditional methods, this classification emerges from optimizing the registration task, ensuring weak smoothing near anatomical boundaries and stronger smoothing in homogeneous regions for more realistic deformations. An interesting strategy was finally presented in \cite{LIU2025103351}, where the authors used spatially-varying losses to favor rigidity constraints on bony structures. In such structures, the authors penalized first and second spatial derivatives of the deformations, with particularly convincing results. We directly extend this work with constraints motivated by solid-mechanics models and applications to local rigidity and sliding motion.

Note that further guidance can be implicitly used during regularization to enforce organ-specific knowledge. As previously stated, segmentation maps can be used in a semi-supervised approach to enforce certain tissue alignments. In that way, anatomy can indirectly guide regularization.

\subsubsection{Hard Constraints in DIR: encoding physical laws}

Optimizing DIR using regularization losses alone often fails to produce anatomically realistic deformations at inference, indicating a somewhat complicated generalization process at times. Beyond the guidance of loss terms, implicit regularization strategies enforce desirable properties on deformation fields by introducing particular transformation models or optimization paradigms which guarantee desirable properties to the deformation field such as smoothness or invertibility.
One of the most widespread approaches concerns enforcing diffeomorphic registration implicitly. Given that the space of diffeomorphisms forms a Riemannian manifold, then integrating a smooth velocity field over time generates a flow of transformations which remain smooth and invertible, ensuring a topology-preserving deformation. \cite{arsigny_log_euclidean} optimize smooth stationary velocity fields (SVF) (or time-dependent velocity fields in the large deformation diffeomorphic metric mapping (LDDMM) \cite{beg2005computing,ASHBURNER2011954}). These velocity fields are then integrated over time to produce smooth and invertible displacement fields. Scaling and squaring is typically used to facilitate the integration step \cite{arsigny_log_euclidean}.
In DLIR, certain multi-level network architectures help produce smooth and accurate large diffeomorphic deformations, for example in \cite{mok_large_2020} a pyramid of networks is employed to predict deformations at different resolutions in a coarse-to-fine fashion.

Finally, beyond modifying the transformation model, several approaches use explicit regularization losses which are computed or enforced using auxiliary networks. In \cite{qin2020biomechanics} a VAE is implicitly trained to reconstruct a class of deformation fields which satisfy certain physical properties (e.g: solution to a given plane-strain problem). On the grounds that once trained, the VAE's latent-space approximates the manifold of biomechanically plausible deformations, it's reconstruction loss acts as a regularizer for a network in a typical DLIR problem.
DIR has also benefited from hyper-network approaches. Hyper-networks are networks trained to produce weights of a secondary network which is then used to solve the main task. For instance, as stated in the previous section, in \cite{hoopes2021hypermorph} an auxiliary network conditioned on the regularization level $\lambda$ produces a DIR network which produces deformation fields satisfying the desired regularization. 

\subsection{Physics-informed DNNs in other applications}\label{ssec:PINNS_others}

Different kind of methods were recently proposed to predict the evolution of physical/mechanical systems with data-driven models, and more specifically using neural-network models.  The early works of \cite{ShiEtAl_NeurIPS2015} have shown that specific neural-network architectures were efficient to predict complex physical phenomena such as precipitation forecasting. Later, \cite{Weinan_CMS_2017,Chen_NeurIPS18} clearly formalized the idea that specific residual DNNs could be seen as discretizations of ordinary differential equations. 
Another approach also consists in incorporating physically relevant layers in the DNN \cite{Bezenac_2019,Zongyi_ICLR21,Takeishi_NeurIPS_2021,Thoreau_2023}. For example, in \cite{Bezenac_2019}, the authors used layers that are suitable for advection-diffusion equations to predict the sea surface temperature.

Closer to our work, the popular paper of \cite{Raissi_JCP_2018} proposed to incorporate underlying physical laws in DNNs by using PDE-based losses. Physics-informed losses first penalize deviations from known physical relations between temporal and spatial derivatives of spatio-temporal data at different points. These losses then favor the learning of prediction rules that respect physical constraints in the hidden layers of the DNN, by using back-propagation \cite{LeCun_Nature_2015}. Although this strategy was a breakthrough in the machine learning literature, the trained predictions of a DNN trained using \cite{Raissi_JCP_2018} remain limited to the interpolation and extrapolation of a dynamic phenomenon with specific initial conditions. For DLIR, the contribution of \cite{Raissi_JCP_2018} however suggests that using losses that take into account the spatial derivatives of the deformations may be an interesting option to favor biomechanically relevant deformations.

\section{Rigid mechanics tools to model local deformations}
\label{sec:rigid_mec_tools}

In order to fully understand how to physically constrain registration mappings to favor physically relevant deformations in DNN-based image registration, we first describe the rigid mechanics tools we will use. Note that we focus on 3D images in this paper, but all notions hold for 2D images. 


\subsection{General notations}\label{ssec:VoxReso}

Let $I_F$ be the fixed image on which a moving image $I_M$ is registered. We denote $\Omega \subset \mathbb{R}^3$ the image domain of $I_F$. 
For readability purposes, the equations given in the main manuscript are in voxel coordinates. 
\ref{app:mm_cpt} develops how to account for millimeter coordinates for paper completeness.



 Local deformations are usually modeled by a displacement field $u: \Omega \mapsto \mathbb{R}^3$ in medical image registration. This means that $x+u(x)$ is the voxel coordinate of $I_M$ mapped to the voxel coordinate $x \in \Omega$ of $I_F$. Another representation of the deformations is the mapping $\Phi (x)= x+u(x)$, which directly encodes the coordinates of the corresponding voxel locations. Their detailed representations are:
\begin{align*}
\label{eq:mapping}
u(x)= 
\begin{pmatrix}
u_1 \left((x_1,x_2,x_3) \right)\\
u_2 \left((x_1,x_2,x_3) \right)\\
u_3 \left((x_1,x_2,x_3 \right))
\end{pmatrix}  \textrm{ and } \\
\Phi(x)=x+u(x) = 
\begin{pmatrix}
\Phi_1 \left((x_1,x_2,x_3) \right)\\
\Phi_2 \left((x_1,x_2,x_3) \right)\\
\Phi_3 \left((x_1,x_2,x_3 \right))
\end{pmatrix} \,.
\end{align*}

\subsection{Mapping Jacobians}\label{ssec:MappingJacobians}

\begin{figure*}[t]
\centering
\includegraphics[width=1.0\textwidth]{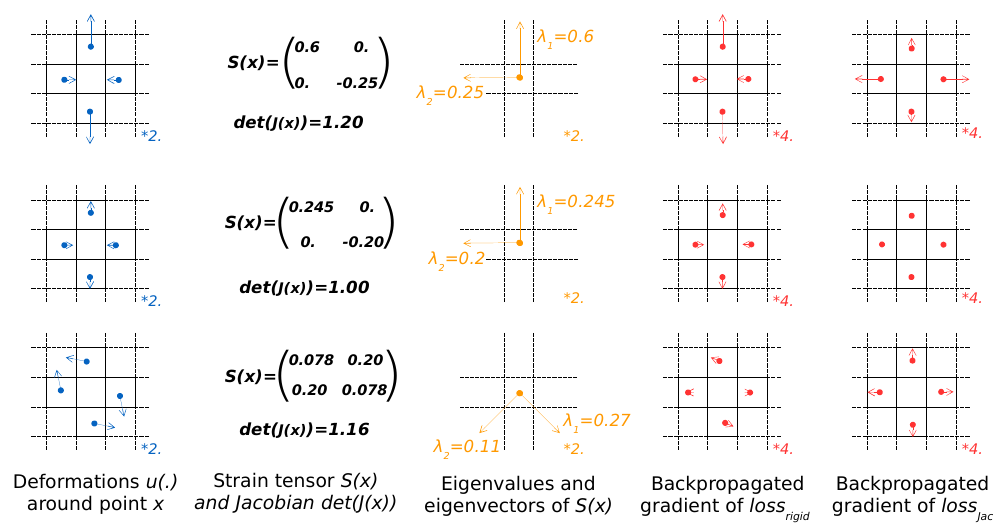}
\caption{Measure  of local deformation   properties and back-propagation of their impact to favor specific bio-mechanical deformation properties using DNN-based image registration. The deformations in this figure are in 2D but the principles are the same in 3D domains. Remark that each represented vector is scaled using the multiplicative factor given at the bottom-right of each sub-figure to be visible. \textbf{(Top)} Strong expansion in one direction and small compression in the orthogonal direction. \textbf{(Middle)} Similar configuration than in the top row, but the local volume property is now preserved, meaning that $\log(det(J(x)))\approx 0$. As a consequence $loss_{Jac}$ as no regularization effect while $loss_{rigid}$ still favors locally rigid deformations \textbf{(Bottom)} Compressions and expansions in diagonal directions. These directions are properly captured by the eigenvectors of the strain tensor $S(x)$ and the corresponding eigenvalues reflect the levels of compression and expanion.}\label{fig:defgrids}
\end{figure*}

A widely used mathematical tool to model the local expansion or compression of deformed structures is the Jacobian matrix of $\Phi$. In voxel coordinates it is:
\begin{align*}
\label{eq:jacobian_vox}
J(x)
&= 
\begin{pmatrix}
\nabla \Phi_1 (x)\\
\nabla \Phi_2 (x)\\
\nabla \Phi_3 (x)
\end{pmatrix} \\
&=
\mathbb{I}_3 +
\begin{pmatrix}
\nabla u_1 (x)\\
\nabla u_2 (x)\\
\nabla u_3 (x)
\end{pmatrix} \\
&=
\begin{pmatrix}
1+\frac{\partial u_1 (x)}{\partial x_1} &\frac{\partial u_1 (x)}{\partial x_2} &\frac{\partial u_1 (x)}{\partial x_3} \\
\frac{\partial u_2 (x)}{\partial x_1} &1+\frac{\partial u_2 (x)}{\partial x_2} &\frac{\partial u_2 (x)}{\partial x_3}\\
\frac{\partial u_3 (x)}{\partial x_1} &\frac{\partial u_3 (x)}{\partial x_2} &1+\frac{\partial u_3 (x)}{\partial x_3}
\end{pmatrix} \,,
\end{align*}
where $\mathbb{I}_3 $ is the identity matrix of size $3 \times 3$. Note that the corresponding Jacobian matrix in millimeter coordinates is given Eq.~\eqref{eq:jacobian_mm}. 
It can be remarked that this matrix is identity  if there are no variations of displacement around $x$. A common tool to model local volume expansion or compression in medical image registration is the determinant of this Jacobian matrix  $det(J(x))$, also known as the Jacobian. Its value is equal to $1$ if the volumes are locally preserved around $x$. As we can see in Fig.~\ref{fig:defgrids}, the  case $det(J(x))=1$ however does not mean that the deformations are locally rigid. The deformed domain may indeed be compressed in one direction and expanded in an orthogonal  direction, so that the local volumes are locally preserved, although the deformations are not locally rigid. 

Note that it is often interesting to consider the logarithm of $det(J(x))$ instead of $det(J(x))$ directly to model local deformation variations. The value of $\log(det(J(x)))$ is indeed $0$ if there is locally no deformation variation. It additionally  varies symmetrically for similar levels compression or expansion, while $det(J(x)$ has values in $]0,1[$ for local compressions and values in  $]1,+\infty[$ for local expansions. 
Let us deepen this claim by considering the physical meaning of the Jacobian: it represents the transformation of one volume unit after deformation. For example, if the voxel resolution is $1mm^3$, then $det(J(x))$ represents the volume of this pixel after deformation. Now, if the voxel volume is divided or multiplied by $2$, then $det(J(x))=0.5$ or $det(J(x))=2$. The variations of $det(J(x))$ compared with no deformation is then $0.5$ or $1.$. Interestingly corresponding $\log{(det(J(x)))}$ are $-0.69$ and $0.69$, which is symmetric around $0$ and therefore appears as more pertinent to homogeneously penalize local compressions and expansions.

\subsection{Strain tensors and their eigendecompositions}\label{ssec:StrainTensors}

 Another tool from solid mechanics to quantify local deformations is strain tensors and its eigendecomposition. Strain tensors are symmetric matrices of dimension $d \times d$, where $d$ is the domain dimension (here $d=3$). In voxel coordinates, the strain tensor of the deformations around point $x$ is:

\begin{align*}
\label{eq:straintensors}
\begin{split}
S(x)&= 
\frac{1}{2}\left[
\begin{pmatrix}
\nabla u_1 (x)\\
\nabla u_2 (x)\\
\nabla u_3 (x)
\end{pmatrix}
+
\begin{pmatrix}
\nabla u_1 (x)\\
\nabla u_2 (x)\\
\nabla u_3 (x)
\end{pmatrix}^{\intercal}
\right]\\
&= \Resize{\columnwidth}{\begin{pmatrix}
\frac{\partial u_1 (x)}{\partial x_1}   & \frac{1}{2} \left( \frac{\partial u_1 (x)}{\partial x_2} + \frac{\partial u_2 (x)}{\partial x_1} \right) & \frac{1}{2} \left( \frac{\partial u_1 (x)}{\partial x_3} + \frac{\partial u_3 (x)}{\partial x_1} \right) \\
\frac{1}{2} \left( \frac{\partial u_2 (x)}{\partial x_1} + \frac{\partial u_1 (x)}{\partial x_2} \right)  & \frac{\partial u_2 (x)}{\partial x_2} & \frac{1}{2} \left( \frac{\partial u_2 (x)}{\partial x_3} + \frac{\partial u_3 (x)}{\partial x_2} \right)\\
\frac{1}{2} \left( \frac{\partial u_3 (x)}{\partial x_1} + \frac{\partial u_1 (x)}{\partial x_3} \right)   & \frac{1}{2} \left( \frac{\partial u_3 (x)}{\partial x_2} + \frac{\partial u_2 (x)}{\partial x_3} \right) & \frac{\partial u_3 (x)}{\partial x_3}
\end{pmatrix}}  \,,
\end{split}
\end{align*}

and corresponding formulation in millimeter coordinates is given in  Eq.~\eqref{eq:strain_tensors_mm}. 
This tensor measures locally the relative changes of positions of points within a body that has undergone deformation. It can be simply estimated using centered differences at each voxel location $x$ of the displacement field $u$. 

Very interestingly for us the eigenvectors $(v_1,v_2,v_3)$ of $S(x)$ are the orthonormal vectors  representing main local directions of expansion or compression of the displacement field $u$. As illustrated Fig.~\ref{fig:defgrids}, the corresponding  eigenvalues $(\lambda_1,\lambda_2,\lambda_3)$ represent the level of expansion or compression in these  directions.  
\begin{align*}
\label{eq:eigen_decomp}
S(x) &= 
\begin{pmatrix}
\vertbar&\vertbar&\vertbar\\
v_1^{\intercal}&v_2^{\intercal}&v_3^{\intercal}\\
\vertbar&\vertbar&\vertbar
\end{pmatrix}
\begin{pmatrix}
\lambda_1&0&0\\
0&\lambda_2&0\\
0&0&\lambda_3
\end{pmatrix}
\begin{pmatrix}
\horzbar&v_1&\horzbar\\
\horzbar&v_2&\horzbar\\
\horzbar&v_3&\horzbar
\end{pmatrix} \\
&= \sum_{i=1}^3 \lambda_i v_i^{\intercal} v_i \,.
\end{align*}
We will use this decomposition of the strain tensors to mechanically constrain the registration deformations in locally rigid structures. 

\subsection{Projection of displacement vectors on a direction} 

Now consider a vector $n$, which represents a direction on which the displacements vectors $u(x)$ should be projected. In our image registration context, this will be useful to locally penalize the variation of deformations in a specific direction only. The projection of a displacement vector $u(x)$ in direction $n$ is then:
\begin{equation}\label{eq:vec_proj}
u_{proj}(x) = \frac{\langle u(x) , n \rangle}{||n||_2}  n  \,,
\end{equation}
where $\langle . , . \rangle$ is the standard scalar product. Note that the displacements of $u(x)$ on the orthogonal plane to $n$ can also simply be computed as $u_{orth}(x)=u(x)-u_{proj}(x)$.


\section{Incorporating 3D mechanical constraints in DNN-based image registration}
\label{sec:Mec_losses}

We now use the mechanical models of Section~\ref{sec:rigid_mec_tools} to constrain the deformations of DNN-based image registration algorithms. As in the PINNs paradigm  \cite{Raissi_JCP_2018}, we constrain the neural-network outputs with physically informed losses. Our losses however only take into account spatial deformations and are applied to displacement fields. Before describing our mechanically-informed losses, we want to emphasize that all the models described in Section~\ref{sec:rigid_mec_tools} can be computed and back-propagated using an automatic differentiation tool like PyTorch for instance. This therefore makes simple their integration in DNN-based image registration tools like VoxelMorph \cite{balakrishnan_unsupervised_2018} or others. 

As in \cite{LIU2025103351}, the deformations around each voxel $x \in \Omega$ are spatially regularized with a loss that depends on local structures properties. We will distinguish locally rigid structures, sliding structures and generic smoothly deforming structures in Subsections~\ref{sec:rigid_losses} to \ref{sec:jacobian_losses}. As these losses are expected to be simultaneously used at all voxel locations, depending on the locally deformed structures, we will also discuss how to weight them in Subsection~\ref{sec:loss_weights}.

\subsection{Local rigidity loss}\label{sec:rigid_losses}

To favor locally rigid  deformations $u(x)$, we first compute the local strain tensor $S(x)$ using Eq.~\eqref{eq:straintensors} and then compute its eigenvalues $\left(\lambda_1(x),\lambda_2(x),\lambda_3(x)\right)$ as in Eq.~\eqref{eq:eigen_decomp}. 
We discussed in Subsection~\ref{ssec:StrainTensors} that the local expansion or compression around $x$ and in all possible directions are limited if these $\lambda_.(x)$ are close to $0$. We then define  the \textit{local rigidity loss} as:
\begin{equation}\label{eq:rigidloss}
loss_{rigid} (x) = \sum_{i=1}^3 \left( \lambda_i(x) \right)^2 \,.
\end{equation}

Remark that most medical image registration methods that rigidly deform specific shapes explicitly use the rotation center in their deformation model, which is not the case for us. We explain in~\ref{app:rigidloss} why the loss  Eq.~\eqref{eq:rigidloss} can favor rigid deformations of whole shapes without taking into account the rotation center.

\subsection{Shearing loss}\label{sec:shear_losses}

The second case we address is more complex than local rigidity as it deals with shearing phenomena, also known in medical imaging as sliding motion. An example of anatomical region where sliding motion occurs is the thoracic cage boundary, next to the lung/liver interface, as illustrated Fig.~\ref{fig:motivating_example}-(right). A large up and down motion occurs there at the lung/liver interface (bottom of the black structure in Fig.~\ref{fig:motivating_example}-(right))  due to breathing, while almost no up and down motion occurs at the level of the ribs (white structures in Fig.~\ref{fig:motivating_example}-(right)). The location of sliding constraint also slightly move in a direction globally orthogonal to the main lung/liver interface motion due to the general swelling of the rib cage during breathing.

When designing a \textit{shearing loss} at a point $x \in \Omega$, where shearing deformations occur, we then want to locally favor smooth deformations in one direction  $n$ but do not want to constrain the deformation smoothness on its orthogonal plane.
We then compute the projections $u^{n}(x_{ngb})$ of $u(x_{ngb})$ on $n$ using Eq.~\eqref{eq:vec_proj}, where the $x_{ngb}$ are the 6-neighbors of $x$. Computing the $x_{ngb}$ in the 6-neighbors of $x$  makes it possible to compute the spatial derivatives of  $u^{n}(x)$ using centered finite differences and then the strain tensor $S^{n}(x)$ of the projected vectors at point $x$ (see Eq.~\eqref{eq:straintensors}). We finally compute its eigenvalues $\left(\lambda_1^{n}(x),\lambda_2^{n}(x),\lambda_3^{n}(x)\right)$ as in Eq.~\eqref{eq:eigen_decomp}. The shearing loss at point $x$ and normal direction $n$ is then
\begin{equation}\label{eq:sheagingloss}
loss_{shearing} (x,n) = \sum_{i=1}^3 \left( \lambda_i^n(x) \right)^2 \,.
\end{equation}
Remark that the same loss can be considered on the residual vector $u(x) - u^{n}(x)$ if one wants to only constraint the  deformations orthogonal to $n$.

\subsection{Jacobian loss}\label{sec:jacobian_losses}

Jacobian of local deformations are commonly used to evaluate local levels of compression or expansion in medical image registration. In our results, we will then use them as a default regularizer when no prior information is known about local deformation properties. We then present in the paragraph below how to define a loss from $det(J(x))$. Importantly, a discussion is also given in \ref{sec:loss_weights} to explain how to simultaneously weight a Jacobian loss at most voxel locations, and weight rigidity or shearing losses at other voxel locations. The conclusion of this discussion is that it makes sense to weight the losses $loss_{rigid}$, $loss_{shearing}$, and $loss_{Jac}$ with the same weights when reasonably smooth deformations are expected. 


In order to define a \textit{Jacobian loss}, we first refer to the discussion Section~\ref{ssec:MappingJacobians}, where we explained why it was interesting to consider the logarithm of $det(J(x))$ instead of $det(J(x))$ directly. Its values are indeed close to $0$ if there is little local compression or expansion. We also claimed that $\log{(det(J(x)))}$ varies symmetrically for similar levels of compression and expansion, contrary to $det(J(x))$. These properties are particularly interesting for a spatial regularization loss, so we propose to use the following \textit{Jacobian loss} to spatially constrain deformations:
\begin{equation}\label{eq:Jacobianloss}
loss_{Jac} (x) =  \left( \log{(det(J(x)))} \right)^2 \,.
\end{equation}
Note that a positive threshold $\epsilon$, with a value close to $0$, is in practice applied to $det(J(x))$ to address potential negative Jacobians.

\section{Identifying Rigid and Shear-Susceptible Regions in Medical Images}
\label{sec:annot_images}

Annotations reflecting the different biomechanical properties of the tissues represented in the training images are required to define the locations of the losses constructed in Section~\ref{sec:Mec_losses}. We hereby describe our protocol to automatically derive these subregions using segmentation masks. Specifically, we used TotalSegmentator \cite{isensee_nnu-net_2021,wasserthal_segmentatorCT,akinci_segmentatorMR} in our work to obtain more than 100 labels associated with each image. Note that  alternative open-source medical image segmentation networks could have been used to obtain detailed segmentation maps. Most public registration datasets also provide segmentation masks, which render our approach universally applicable.

\begin{figure*}
    \centering
    \includegraphics[width=\textwidth]{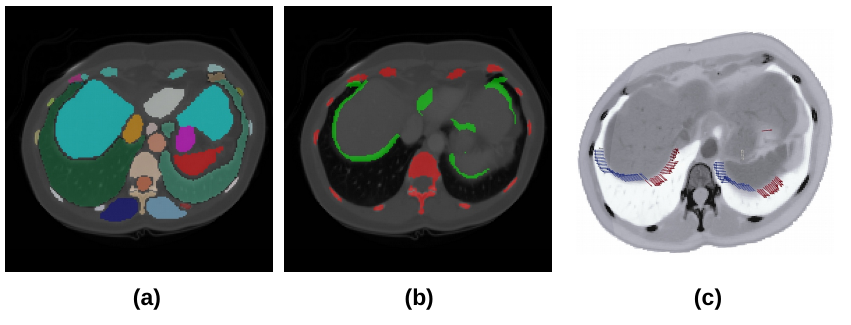}
    \caption{Identification of the regions with locally-rigid or sliding deformations in a 3D volume out the \textit{Learn2Reg AbdomenCTCT} dataset \cite{Zhoubing_learn2reg}.
    \textbf{(a)} Segmentation labels automatically generated. 
    \textbf{(b)} Derived regularization regions ($\mathcal{R}$) for rigidity,shearing and Jacobian losses.
    \textbf{(c)} Subset of projection vectors $\mathit{Strain\_Directions}$ derived for the shearing loss.}
    \label{fig:sampleAbdomenCTCT}
\end{figure*}

\subsection{Regularization Mask}\label{ssec:reg_mask}

In order to correctly regularize the deformations of different anatomical structures, we construct a regularization mask $\mathcal{R} \in \{R,S,J\}^{H,W,D}$ where $H,W,D$ refer to the voxels in the image domain $\Omega$, and  $\{R,S,J\}$ are the labels indicating the location of local rigidity, shearing and Jacobian losses, respectively. An illustration of $\mathcal{R}$ is given in Fig.~\ref{fig:sampleAbdomenCTCT}(b). More details about the tissues and organ interfaces we used to construct $\mathcal{R}$ are also developed in \ref{app:reg_regions}.

\paragraph{Rigidity constraints $R$}
To begin with, the rigidity loss Eq.~\eqref{eq:rigidloss} characterized by near-zero local contractions and expansions was assigned to hard tissues. These typically include bones (\textit{e.g.} vertebrae, ribs, sternum and costal cartilages in abdominal images). 

\paragraph{Sliding motion $S$}
Likewise, we use a list of tissue pairs between which sliding motions (\textit{i.e}. shearing forces) are susceptible to arise. For each pair of these tissues, we extract the overlapping regions between them after dilating the associated masks to cover a larger area. We assign the shearing regularization label $S$ to these regions and use the loss Eq.~\eqref{eq:sheagingloss} there. 

\paragraph{Default Jacobian constraint $J$}
We finally assign a Jacobian loss  Eq.~\eqref{eq:Jacobianloss} to other voxels. This favors pseudo-elastic deformations there.

\subsection{Shearing Loss Projection Directions}\label{ssec:reg_directions}

As described Section~\ref{sec:shear_losses}, the shearing loss makes use of local normal vectors $n$ to the sliding surface. 
To compute these vectors, we consider the voxels $x \in \mathbb{S}$ such that $\mathbb{S} = \{ x \, \lvert \,  \mathcal{R}(x) = S \}$. Given that $\mathbb{S}$ consists of distinct regions with predominantly flat surfaces, normal directions at these points can be estimated as follows: for each voxel $x \in \mathbb{S}$, we identify its $N$ nearest neighbors within $\mathbb{S}$ (with $N=20$ in our experiments). The resulting point cloud is then analyzed using principal component analysis (PCA), which decomposes it into three principal directions. The first two components describe the tangent directions of the shearing region, while the third component provides the normal direction. At each voxel $x \in \mathbb{S}$, we then normalize this third component and get $n(x)$. 

As illustrated Fig.~\ref{fig:sampleAbdomenCTCT}(c), the vectors $n(x), \, x \in \mathbb{S}$ are finally stored in a tensor $\mathit{Strain\_Directions} \in \mathcal{R}^{(3,H,W,D)}$, where the voxels $x \notin  \mathbb{S}$ contain null displacements.


\section{Materials and Methods}\label{sec:materials_methods}

Before presenting and discussing our results Section~\ref{sec:Results}, we describe hereafter our experimental protocol. Our strategy will first be assessed on 3D synthetic data designed to showcase controlled rigid and shearing body motions.  We will then evaluate its generalization properties on real 3D data out of the \textit{Learn2Reg AbdomenCTCT} challenge \cite{Zhoubing_learn2reg}. All traininng scripts are available on github \footnote{\url{https://github.com/Kheil-Z/biomechanical_DLIR}}. 

\subsection{Synthetic Data} \label{ssec:MatSyntheticData}

\paragraph{Motivation}
In order to validate and quantify our proposed regularization losses in a controlled setting, we generate two synthetic toy datasets tailored to the specifics of our task. Both datasets consist of volumetric images of size (64,64,64) featuring rigid cuboids to which we instill known mechanical constraints. To investigate how well the trained networks generalize the learned deformation properties at inference,  these cuboids are generated with random sizes, center positions, and movement amplitudes such that a large panel of various images are available, with differences between training, validation sets and test sets. We also keep track of the cuboid locations by saving their segmentation masks. 

\begin{figure}[htb]
         \centering
         \includegraphics[width=1.0\linewidth]{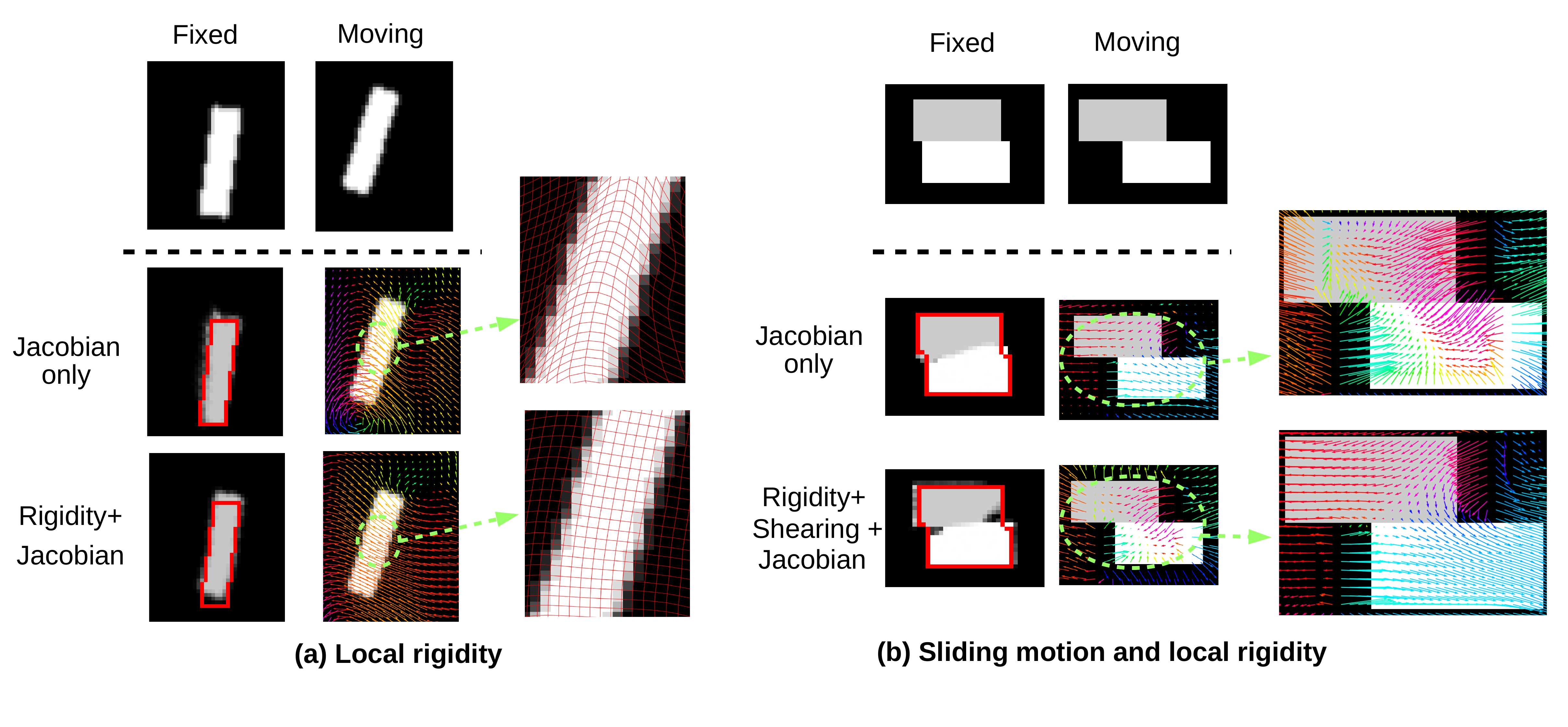}

        \caption{Sample registration results from \textbf{(a)} $\mathcal{D}_{\textit{rigid}}$ and  \textbf{(b)} $\mathcal{D}_{\textit{shearing}}$ test sets. In both cases, the first row displays the fixed and moving images. The second and third rows present the moved images on the left with the fixed target's outline, while the right side shows the deformation fields overlaid on the moving images. Notably, the rigid deformations inside the cuboids are nearly rigid, while the sliding effect is well generalized to test data.}
        \label{fig:sample_rota_shear_registration}
\end{figure}

\paragraph{Rigid Dataset $\mathcal{D}_{\textit{rigid}}$} The first dataset, $\mathcal{D}_{\textit{rigid}}$ illustrated Fig.~\ref{fig:sample_rota_shear_registration}(a), allows to examine the impact of the rigidity loss term Eq.~\eqref{eq:rigidloss}. We generate single cuboids in every image, and randomly rotate and translate them. Thus, taking two randomly generated images entails the registration problem, where a rigid object is rotating. An optimal deformation field should hence respect the rigidity constraint inside the cuboid voxels, while rotating the entire object. 

\paragraph{Shearing Dataset $\mathcal{D}_{\textit{shearing}}$} The second dataset, $\mathcal{D}_{\textit{shearing}}$, illustrated Fig.~\ref{fig:sample_rota_shear_registration}(b), is designed to observe sliding motions between objects. For every image, we generate two distinct but bordering cuboids, then translate them in opposing directions so as to observe a shearing region on their adjoining borders. Note that the interface location between the objects is also randomly drawn around the image center. The registration problem between a moving and fixed image then attempts to obtain the relative translation between the corresponding positions of the objects. Associated segmentation masks allow us to enforce a rigid loss inside the distinct objects, and permit a shearing region on their surface of contact.

\paragraph{Cross-validation strategy}
In both cases, 200 pairs of images are used for training, 50 for validation and 50 for testing. We recall that the segmentation masks are only used when registering the training images. 
Note again that in both cases, once trained, the model only infers deformation fields using the paired fixed and moving images from the unseen test set, with no notion of segmentation masks or various regularization regions other than the deduced correlations from the training step.

\subsection{Real Data}
\label{ssec:MatRealData}

\begin{figure}[htb]
         \centering
        \includegraphics[width=1.0\linewidth]{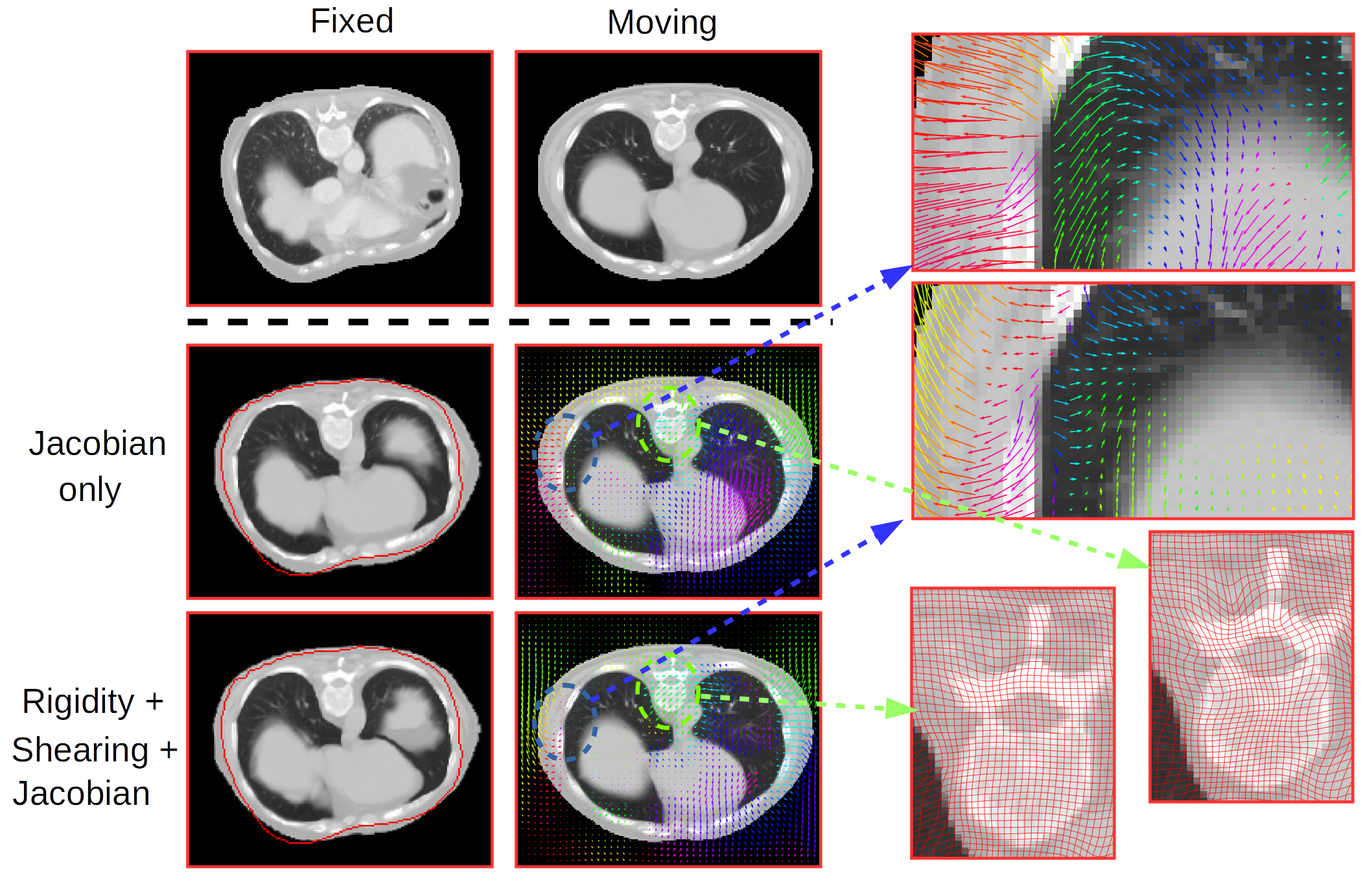}

        \caption{Sample registration results of a patient from the AbdomenCTCT dataset. The first row contains the fixed and moving images and the next rows display the moved image with an outline of the desired result, and the associated predicted deformation fields plotted over the moving images. We also display two regions of the obtained deformation field. }
        \label{fig:sample_abdomenctct_registration}
\end{figure}

\paragraph{Motivation} In view of demonstrating the real-life applicability of the proposed approach, we further assess it on 3D medical images out of the well established  \textit{Learn2Reg challenge}\footnote{Learn2Reg challenge: \href{https://learn2reg.grand-challenge.org/Description/}{https://learn2reg.grand-challenge.org/Description/}}.
An illustration of registered images is given Fig.~\ref{fig:sample_abdomenctct_registration}.

\paragraph{Learn2Reg AbdomenCTCT dataset} We specifically trained our networks on the \textit{AbdomenCTCT} dataset, which contains 30 unpaired (inter-patient) Abdominal CT scans resize to a fixed resolution of $(192,160,256)$ voxels. Alongside the images, the challenge provides annotations of $13$ anatomical labels. We choose to work on this dataset due to its intrinsic challenges: large deformations alongside a limited number of images. The dataset also features the complete abdominal region (lungs to pelvis), allowing us to incorporate several regions of rigid and shearing constraints. As a standard pre-processing step, we first down-sample the volumes by a factor $2$, clip the voxels intensity range to $[-1200, 400]$ Hounsfield units, and min-max normalize them to $[0,1]$. Finally, as specified in Section~\ref{sec:annot_images}, we add annotation labels relative to vertebrae, ribs and lung nodules to the training segmentation masks provided by the challenge. 

\paragraph{Cross-validation strategy} We employ the \textit{Learn2Reg} predefined image pairs for validation (45 pairs), meanwhile we train our networks by randomly pairing the remaining un-paired images not included in the validation set (190 pairs). 

\subsection{DNN Training}
\label{ssec:MatDNN}

\paragraph{DNN architecture and general training strategy}
 In all our experiments, we trained a Voxelmorph-style registration network using a more modern Attention U-Net \cite{oktay2018attentionunetlearninglook} backbone to register image pairs. The network architecture contains 5 layers of (32, 64, 128, 256, 512) channels. We also consider a diffeomrophic setting, where the network predicts a velocity field, which is then integrated by scaling-and-squaring \cite{arsigny_log_euclidean} over 7 time steps to produce the final deformation field.  The networks are trained for 100 epochs using the Adam optimizer, with a fixed learning rate of $1e-3$ and a batch size of 8. As detailed in Subsections \ref{ssec:MatSyntheticData} and \ref{ssec:MatRealData}, cross-validation was performed by splitting train, test and validation image pairs.

\paragraph{Training loss for the synthetic datasets}
 The following loss was used on  synthetic datasets:
\begin{equation}\label{eq:loss_synthdata}
    \mathcal{L} = \alpha MSE(I_F,I_M \circ \Phi) +\lambda \mathcal{L}_{reg}(\Phi) \,,
\end{equation}
where $\lambda \in [0,1]$, $\alpha=1-\lambda$ and MSE is the mean squared error between the fixed image and the warped moving image. We ran a hyper-parameter grid search on 13 values of $\alpha$ from 0.99 to 0.01.

\paragraph{Training loss for real medical images}
In the case of the real dataset, we include a Dice loss on the warped segmentation mask of the moving image $S_M$ for additional structural awareness:
\begin{equation}\label{eq:loss_realdata}
    \mathcal{L} = \alpha MSE(I_F,I_M \circ \Phi) + \gamma Dice(S_F,S_M \circ \Phi) +\lambda \mathcal{L}_{reg}(\Phi) \,,
\end{equation}
where $\alpha \in [0,1]$, $\gamma \in [0,1]$, $\lambda \in [0,1]$, with the constraint $\alpha+\gamma+\lambda=1$. We performed a grid search to identify the optimal hyper-parameters, varying $\alpha$ between 0.9 and 0.3. For each $\alpha$, we tested values of $\lambda$ and $\gamma$ between 0.1 and $1-\alpha$. In total, 24 combinations of $(\alpha, \lambda, \gamma)$ were evaluated for each loss type.

\paragraph{Compared regularization losses $\mathcal{L}_{reg}$} To ensure a fair comparison and to understand the effects of our proposed loss term, we identically trained the DNN with different hyper-parameters on all datasets. We also compared the effect of the well-established Jacobian loss regularization term in all the data domain to our proposed loss terms. 
Specifically, the regularization loss terms $\mathcal{L}_{reg}$ used to penalize the predicted DDF $\Phi$ throughout experiments are: 
\begin{itemize}
    \item \emph{Jacobian Only}: $\mathcal{L}_{reg}(\Phi) = loss_{Jac}(\Phi)$,
    \item \emph{Rigiditiy + Jacobian}: $\mathcal{L}_{reg}(\Phi) = loss_{rigid}(\Phi \odot \mathbb{R}) + loss_{Jac}(\Phi \odot \mathbb{J})$,
    \item \emph{Rigidity + Shearing + Jacobian}: $\mathcal{L}_{reg}(\Phi) = loss_{rigid}(\Phi \odot \mathbb{R})  + loss_{shearing}(\Phi \odot \mathbb{S},  \mathit{Strain\_Directions}) + loss_{Jac}(\Phi \odot \mathbb{J})$,
\end{itemize}
where $loss_{Jac}, loss_{rigid}, loss_{shearing}$ are defined in Section~\ref{sec:Mec_losses}, $\mathit{Strain\_Directions}$ is the tensor of projection directions defined in Section~\ref{ssec:reg_directions}, and $\odot$ denotes the Hadamard product between the deformation field $\Phi$ and the masks $\mathbb{J}, \mathbb{R}, \mathbb{S}$ of Section~\ref{ssec:reg_mask}.

\subsection{Metrics}
\label{ssec:MatMetrics}

Quantitative evaluation was used to quantify the impact of the proposed loss terms. For each considered hyper-parameters set, the trained networks are independently evaluated on the unseen test sets. 
Note that the qualitative evaluation of the deformations will also be discussed based on different figures.

\paragraph{Registration accuracy}
To quantify registration accuracy, we average the registration MSE error after registration over all test cases. When applicable, we also average the Dice scores for labels provided by the Learn2Reg ($Dice_{learn2reg}$) challenge. Dice scores are also computed on $Dice_{additional\_labels}$ and $Dice_{lung\_nodules}$.

\paragraph{Bio-mecanical properties of the predicted deformation fields}
To describe the physical behavior of the deformation fields $\phi$ using the  deformation Jacobians $\lvert J_{\phi} \rvert$, we first compute the percentage of voxels containing 3D foldings, \textit{i.e.} $\lvert J_{\phi} \rvert \leq 0$ throughout the field, as well as the standard deviation of their logarithm: $SDlog  | J_{\Phi} |$. When studying the shearing prone datasets \textit{AbdomenCTCT} and $\mathcal{D}_{\textit{shearing}}$, this last metric is computed in $\overline{\mathbb{S}} = \mathbb{R} + \mathbb{J}$ in order to quantify the deformation field's smoothness outside of the regions intended to present shearing motion. We denote $ SDlog  | J_{\Phi \odot \overline{\mathbb{S}}} |$ this anatomy-aware metric.
We additionally use the rigidity map $\mathbb{J}$ of Section~\ref{sec:rigid_losses} to compute  $\mathbb{L}_{rigid} := \sum_{i=1}^3 \left( \lambda_i(\Phi \odot \mathbb{J}) \right)^2$. This metric quantifies the deviation of the deformation field from rigid displacements in $\mathbb{J}$, \textit{i.e.} inside the cuboids or the bone structures.


\section{Results and Discussions}
\label{sec:Results}

\begin{figure}[htb]
        \centering
        \includegraphics[width=0.7\columnwidth]{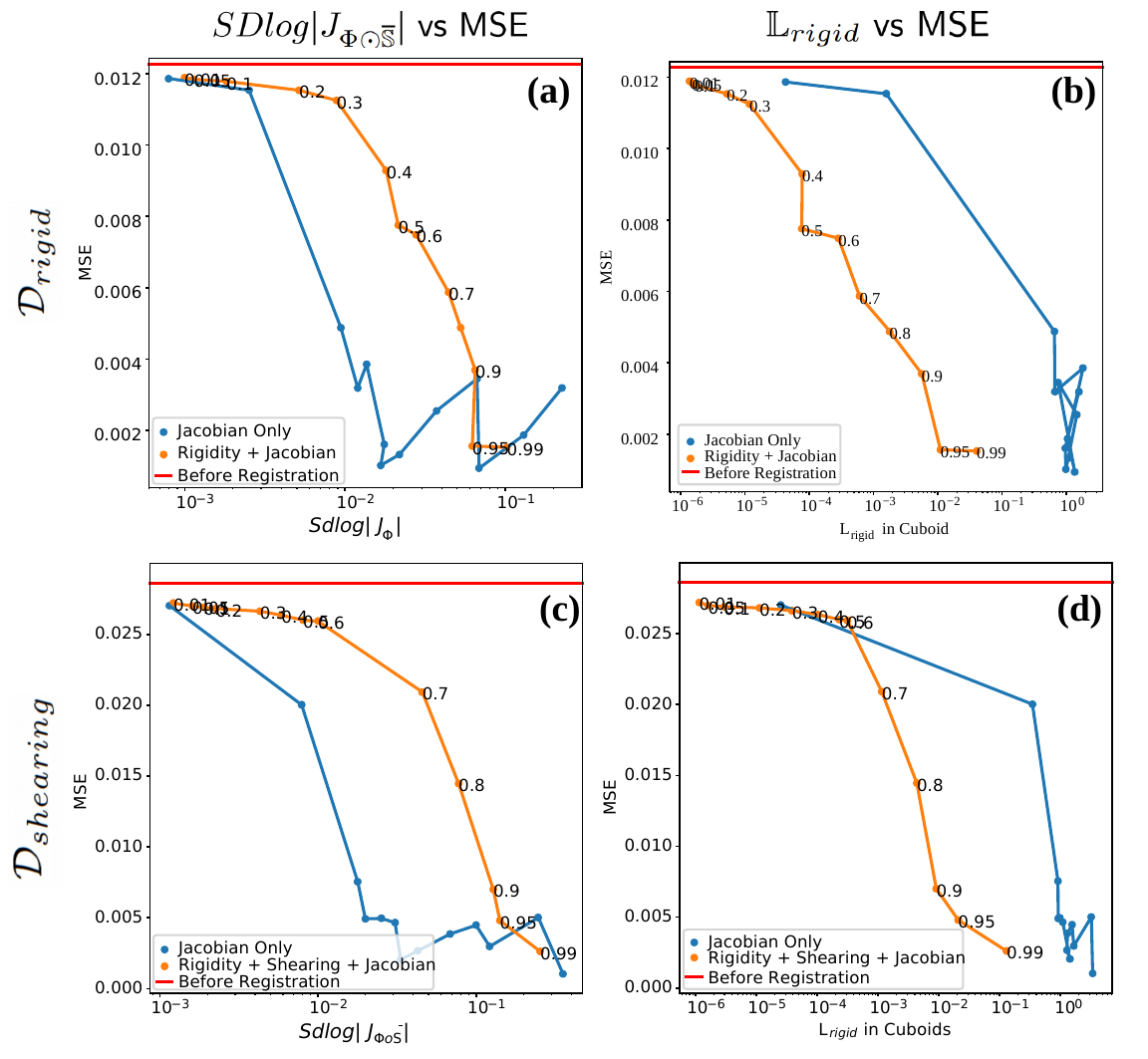}
        \caption{A sensitivity analysis of the effects of loss coefficients on the test sets of simulated datasets. The top row \textbf{(a-b)} corresponds to $\mathcal{D}_{\textit{rigid}}$ while the second row \textbf{(c-d} represents $\mathcal{D}_{\textit{shearing}}$. In the left column \textbf{(a-c)}, $SDlog  | J_{\Phi} |$ is plotted against MSE, whereas the right column \textbf{(b-d)} illustrates the relationship between measured rigidities inside the cuboids and the corresponding MSE measures.
        The plotted values reflect the weighting of the similarity term in the loss function. Blue curves represent models trained exclusively with the Jacobian term, while orange curves correspond to models incorporating our proposed physical regularization term.}
        \label{fig:ablation}
\end{figure}

\subsection{Ablation Study on the synthetic datasets}


We first performed an ablation study to analyze the impact of varying loss term weights. For each considered loss term, we trained 13 models, adjusting the weight ratio $\lambda$ between the MSE loss and the regularization loss. The MSE loss weight varied from $0.99$ to $0.01$, while the regularization loss weight was set to $1-\lambda$.
Figures~\ref{fig:ablation}(a) and \ref{fig:ablation}(c) illustrate how the regularity of the inferred displacement fields, measured by $SDlog  | J_{\Phi} |$ related to the average MSE from the registration. Figures~\ref{fig:ablation}(b) and \ref{fig:ablation}(d) also show how deformation rigidity inside the cuboid compare to the resulting MSE across the image. In both cases, metrics are averaged over the unseen test set.

Across both synthetic datasets and regularization strategies, we observed the expected trade-off between field regularity and over-accurate registration. More interestingly,  we identified a competing balance between overall field smoothness and adherence to physical properties, as evidenced by the alternating optimal fronts between Figures~\ref{fig:ablation}(a) and \ref{fig:ablation}(b), and between Figures~\ref{fig:ablation}(c) and \ref{fig:ablation}(d). Notably, these trade-offs differ in scale: improvements in \textit{Rigidity + Jacobian} and \textit{Rigidity + Shearing + Jacobian} lead to a substantial gain in rigidity, yet at the cost of a slightly less optimal $SDlog  | J_{\Phi} |$, with an order of magnitude separating these respective effects. Moreover, we remark that our physics-inspired losses tend to provide a smoother regularization effect, whereas the Jacobian-only loss appears highly sensitive at high $\alpha$ (i.e., minimal regularization), leading to more abrupt changes in the model’s behavior. These results in all cases suggest that the bio-mechanics losses we used on training data have the desired impact when predicting the deformations on test data.

\subsection{Detailed results on the synthetic dataset $\mathcal{D}_{\textit{rigid}}$}
\label{ssec:ResSyntheticData}


\begin{table*}[htb]
\begin{center}
\caption{$\mathcal{D}_{\textit{rigid}}$ test set results (\textit{mean $\pm$ std}) for different regularization strategies. Best results are in bold, and second best in italics. Note that all regularization strategies yielded strictly 0 \% of folding voxels, meanwhile not applying a regularization resulted in approximately 4.7\% of foldings over the whole deformation.}
\label{tab:rotation_best_mse}
\footnotesize
\begin{tabular}{cllll}
\toprule
Regularization & $ SDlog  | J_{\Phi} | $ $\downarrow$ & $\mathbb{L}_{rigid}$ $\downarrow$ & MSE $\downarrow$ \\ 
\midrule
Jacobian Only &  \textbf{0.0167 $\pm$ 0.0039} & \textit{0.9876 $\pm$ 0.8893} &  0.0010 $\pm$ 0.0008 \\
Rigidity + Jacobian &  \textit{0.0620 $\pm$ 0.0153} &  \textbf{0.0110 $\pm$ 0.0096} &  0.0016 $\pm$ 0.0015 \\
\textit{Before Registration} &                   -- &                   -- &  0.0123 $\pm$ 0.0039 \\
\bottomrule
\end{tabular}
\end{center}
\end{table*}

We now demonstrate on $\mathcal{D}_{\textit{rigid}}$ the efficacy of the proposed rigidity loss 
to produce near-rigid deformations inside a rigid object.
Table~\ref{tab:rotation_best_mse} displays the means and standard deviations of the evaluation metrics computed on the test set of $\mathcal{D}_{\textit{rigid}}$ with parameters $\lambda$ leading to similar MSEs on test data.  

We start by noting the steep improvement (two orders of magnitude) of $\mathbb{L}_{rigid}$ inside the cuboid indicating that the use of our rigidity loss permits to generalize stiffness properties of objects.
Although an increase in $SDlog  | J_{\Phi} |$ is noted, it is negligible as our experiments show that a model with no regularization whatsoever results in a much higher $SDlog  | J_{\Phi} | = 1.0151 \pm 0.0472$. Additionally, in both cases the models produce no foldings.

We conclude that the rigidity loss indeed allows a neural network to produce deformation fields which respect structural rigidity of objects contained in the training set, with no effect on DIR accuracy and the rest of the deformation.  This can be further appreciated on Figure~\ref{fig:sample_rota_shear_registration}(a). 
Although we depict a 2-D slice of the object, we clearly see that the uniformly spaced grid remains unaltered with our proposed loss term.

\begin{table*}[htb]
\begin{center}
\caption{$\mathcal{D}_{\textit{shearing}}$ test set results (\textit{mean $\pm$ std}) for different regularization strategies. Best results are in bold, and second best in italics. Note that all regularization strategies yielded less than 0.5 \% of folding voxels, meanwhile not applying a regularization resulted in approximately 8.8\% of foldings over the whole deformation.}
\label{tab:shearing_best_mse}
\begin{tabular}{llll}
\hline
\multicolumn{1}{c}{Regularization Term} & $ SDlog  | J_{\Phi \odot \overline{\mathbb{S}}} | $ $\downarrow$ & \multicolumn{1}{c}{$\mathbb{L}_{rigid}$ $\downarrow$} & \multicolumn{1}{c}{MSE  $\downarrow$} \\ \hline
Jacobian Only                           & 0.3540 $\pm$ 0.0538                  & 3.3734 $\pm$ 0.5850                                   & \textbf{0.0010 $\pm$ 0.0011}          \\
Rigidity + Shearing + Jacobian          & \textbf{0.2542 $\pm$ 0.0588}         & \textit{\textbf{0.1298 $\pm$ 0.0814}}                 & 0.0026 $\pm$ 0.0017                   \\
Before Registration                     & -                                    & -                                                     & 0.0286 $\pm$ 0.0077                   \\ \hline
\end{tabular}%
\end{center}
\end{table*}

\subsection{Detailed results on the synthetic dataset $\mathcal{D}_{shearing}$}

We now present the effect of the shearing constraint on $\mathcal{D}_{shearing}$. Table~\ref{tab:shearing_best_mse} quantifies the averages and the associated standard deviations of the evaluation metrics computed on the test set of the dataset. Again, we display results for hyper-parameters which minimize the registration loss. Identically to the previous dataset, we note through the sharp decrease in $\mathbb{L}_{rigid}$ (measured only inside $\mathbb{R}$) that the rigid property of the two cuboids is reflected in the produced deformation fields, with the registration MSE nearly unaffected. This time over, the model obtained with the mechanics inspired loss term displays a more regular field (lower $SDlog  | J_{\Phi} |$), although it is also an insignificant improvement as both models display near 0\% folding voxels.

The effects of the shearing loss term on the shearing prone region $\mathbb{S}$ can be fully appreciated in the sample registrations provided in Figure~\ref{fig:sample_rota_shear_registration}(b). In both test set cases shown, but more notably on the large deformation one (right sample), we observe a thin region of discontinuity in the deformation on the interface between the cuboids, which allows the \textit{Rigidity + Shearing + Jacobian} model to better grasp the independent, rigid, sliding motion of the cuboids and produce a field satisfying the objects' physical properties. We can also note the uniformity of the field directions inside the cuboid themselves due to both the shearing loss but also the rigidity constraint learned in the objects.

\subsection{Real 3D volumes}\label{ssec:ResRealData}

\begin{figure}[htb]
         \centering
        \includegraphics[width=1.0\linewidth]{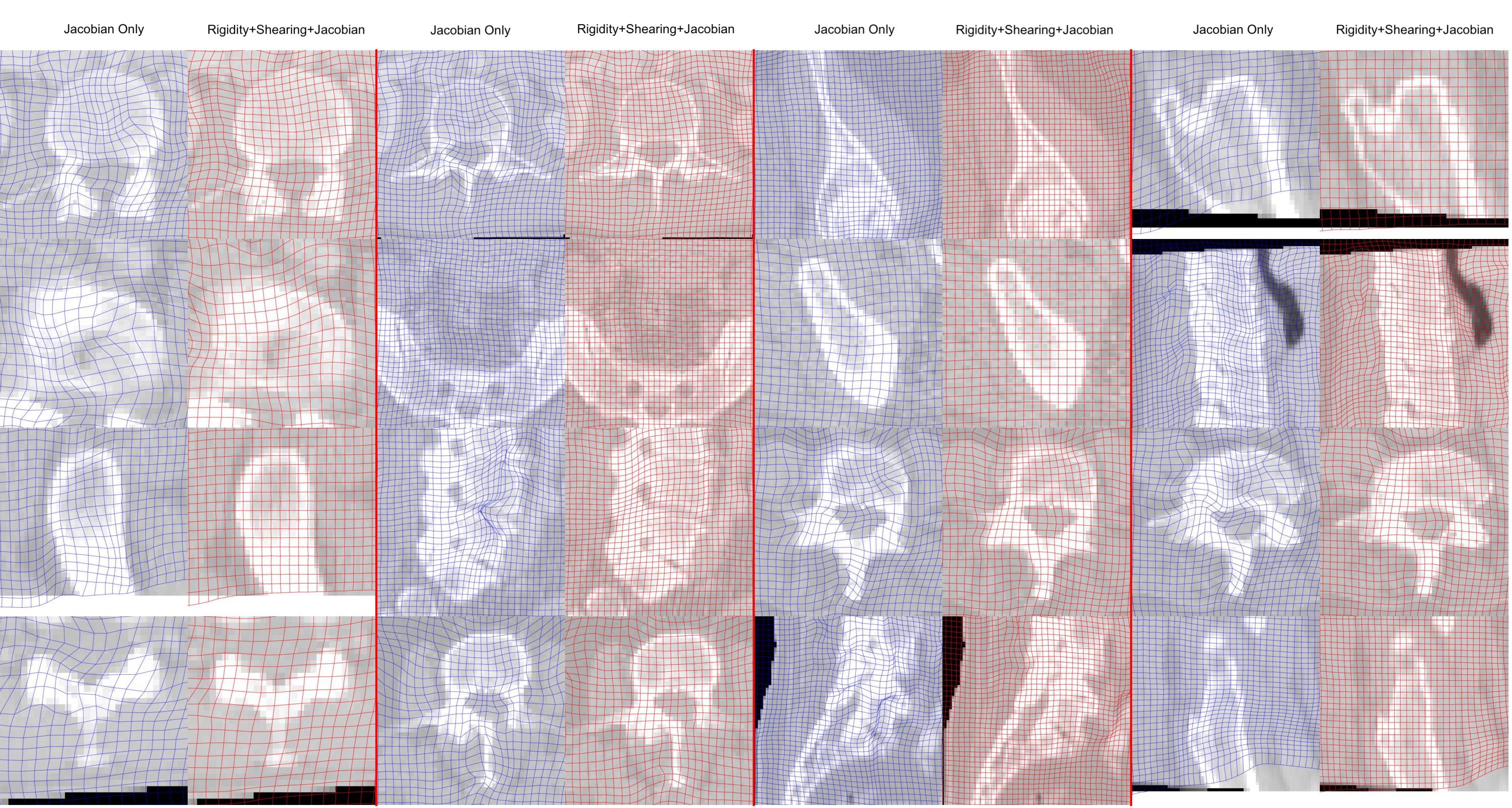}

        \caption{Grid display of deformation fields produced by both models. They are randomly sampled on structures marked as rigid. We overlay the moving images over the deformation fields. The columns alternate between \textit{Jacobian Only} (blue grids) and \textit{Rigidity + Shearing + Jacobian} (red grids). We can clearly note improved rigidity for the second columns.}
        \label{fig:collage_rigid}
\end{figure}

\begin{figure}[htb]
         \centering
        \includegraphics[width=1.0\linewidth]{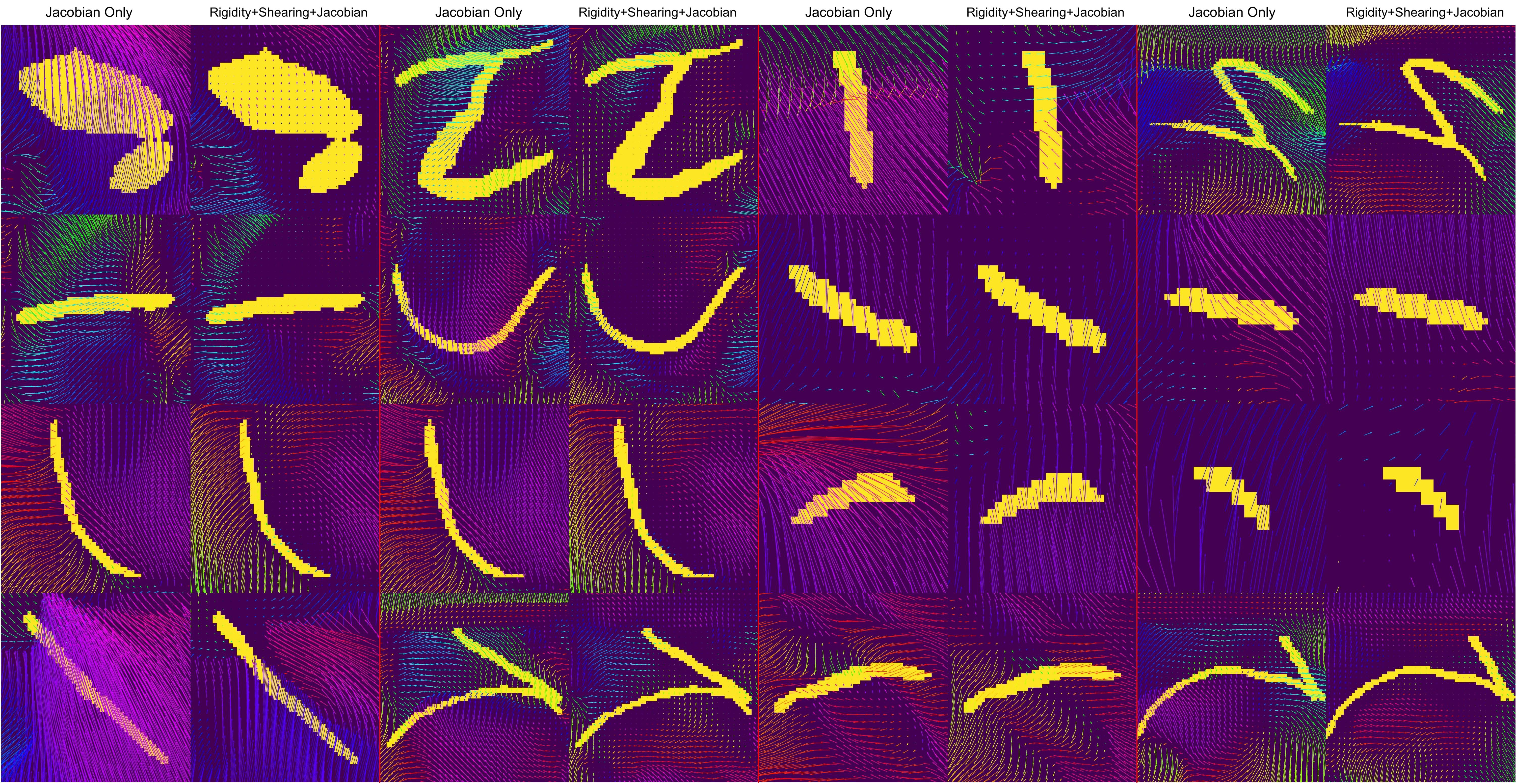}

        \caption{Grid display of deformation fields produced by the two models. They are randomly sampled on regions were the shearing loss was applied in one case. We overlay region of shearing. The columns alternate between \textit{Jacobian Only} and \textit{Rigidity + Shearing + Jacobian}. We can note that in most cases, the model trained with our shearing loss can produce discontinuities inside those regions.}
        \label{fig:collage_shear}
\end{figure}

As described in section~\ref{ssec:MatRealData}, we also evaluated our strategy on real 3D thoracic images presenting major DIR challenges. The training dataset is indeed relatively small (30 unpaired 3D images out of the Learn2Reg challenge) and inter-patient registration is additionally performed. This context allowed us to better grasp the effects of constraining a DNN using intrinsic physical properties of organs.

Table~\ref{tab:abdomenctct_results} synthesizes the results of the different regularization strategies on the AbdomenCTCT dataset of the Learn2Reg challenge. 
We first note that across all experimental settings, our regularization better preserves the rigid structure of hard tissues in the deformation fields. Improvements range from a two fold to a hundred fold reduction in $\mathbb{L}_{rigid}$. Additionally, overall field regularity is better maintained in two out of three settings of Table~\ref{tab:abdomenctct_results}, with the \textit{Rigidity + Shearing + Jacobian} model achieving the lowest $SDlog \lvert J_\Phi\rvert$. 

Crucially, these gains in field regularity and physical plausibility come without sacrificing alignment accuracy. Across MSE, Normalized Cross-Correlation (NCC), and Dice scores, our model performs at least on par with traditional regularization methods. In the medical dataset, it even consistently outperforms them across all accuracy metrics. While the improvements are sometimes small, they remain systematically significant ($p<10^{-3}$ , Wilcoxon signed-rank test). Details on results per organ can be found in Fig.~\ref{fig:detailed_dice} of \ref{app:detailed_additional}.

\begin{table*}[htb]
\caption{AbdomenCTCT validation set results (\textit{mean $\pm$ std}) for different regularization strategies. Best results are in bold, and second best in italics. Note that all regularization strategies yielded less than $3e^{-3}$ \% of folding voxels, meanwhile not applying a regularization resulted in approximately 18.1\% of foldings over the whole deformation. Unless stated otherwise (n.s.), all improvements are statistically significant ($p<10^{-3}$) according to a Wilcoxon signed-rank test.}
\label{tab:abdomenctct_results}
\resizebox{\textwidth}{!}{%
\begin{tabular}{lllllllll}
\cline{2-9}
& Regularization Term & $ SDlog  | J_{\Phi \odot \overline{\mathbb{S}}} | $ $\downarrow$ & \begin{tabular}[c]{@{}l@{}}$\mathbb{L}_{rigid}$ $\downarrow$\\  (Rigidity in Bones)\end{tabular} & MSE $\downarrow$ & NCC $\downarrow$ & $Dice_{learn2reg}$ $\uparrow$ & $Dice_{additional\_labels}$ $\uparrow$ & $Dice_{lung\_nodules}$ $\uparrow$ \\ \cline{2-9} 
\multirow{2}{*}{\textit{Best MSE}} & Jacobian Only & 0.3273 $\pm$ 0.1185 & 0.4454 $\pm$ 0.4572 & \textit{0.0188 $\pm$ 0.0143} & \textit{-0.1889 $\pm$ 0.0213} & 0.2300 $\pm$ 0.0566 & 0.4481 $\pm$ 0.0479 & 0.7139 $\pm$ 0.1274 \\ 
& Rigidity + Shearing + Jacobian & 0.3873 $\pm$ 0.1021 & 0.1374 $\pm$ 0.1956 & \textbf{0.0186 $\pm$ 0.0156} (n.s) & \textbf{-0.1984 $\pm$ 0.0248} & 0.2715 $\pm$ 0.0648 & 0.4732 $\pm$ 0.0530 & 0.7921 $\pm$ 0.1080 \\ \cline{2-9} 
\multirow{2}{*}{\textit{Best Dice}} & Jacobian Only & 0.4239 $\pm$ 0.0419 & 0.8616 $\pm$ 0.2020 & 0.0398 $\pm$ 0.0257 & -0.1783 $\pm$ 0.0207 & \textit{0.3540 $\pm$ 0.0804}  & \textit{0.5074 $\pm$ 0.0598} & \textbf{0.9047 $\pm$ 0.0580} \\
& Rigidity + Shearing + Jacobian & 0.3755 $\pm$ 0.0481 & \textit{0.1043 $\pm$ 0.0397} & 0.0355 $\pm$ 0.0241 & -0.1839 $\pm$ 0.0227 & \textbf{0.3708 $\pm$ 0.0725}  & \textbf{0.5162 $\pm$ 0.0679} & \textit{0.8851 $\pm$ 0.0608} \\
\cline{2-9} 
\multirow{2}{*}{\textit{Best  $ SDlog  | J_{\Phi} | $}} & Jacobian Only& \textit{0.1433 $\pm$ 0.0308} & 1.1492 $\pm$ 0.5114 & 0.0481 $\pm$ 0.0291 & -0.1500 $\pm$ 0.0167 & 0.2957 $\pm$ 0.0637 & 0.4752 $\pm$ 0.0517 & 0.7990 $\pm$ 0.0983 \\
& Rigidity + Shearing + Jacobian & \textbf{0.0966 $\pm$ 0.0165}  & \textbf{0.0130 $\pm$ 0.0130} & 0.0559 $\pm$ 0.0311          & -0.1495 $\pm$ 0.0191 (n.s) & 0.2920 $\pm$ 0.0648 (n.s) & 0.4779 $\pm$ 0.0587 (n.s) & 0.7515 $\pm$ 0.1072 \\ \cline{2-9} 
& Before Registration & - & - & 0.0894 $\pm$ 0.0347 & -0.1122 $\pm$ 0.0156 & 0.2531 $\pm$ 0.0702 & 0.4543 $\pm$ 0.0530 & 0.5939 $\pm$ 0.1163 \\ 
\end{tabular}%
}
\end{table*}

Beyond improving alignment and regularity metrics, visualizing the deformation fields in different regions of the CT scans helps further grasp the extent of the results. We first examine the deformation fields generated within hard tissues for both models. Random samples are shown in Fig.~\ref{fig:collage_rigid}. The regular grid structure is preserved within the bones for the model trained with our loss function. In shearing-prone regions, Fig.~\ref{fig:collage_shear} shows that the \textit{Rigidity + Shearing + Jacobian} model sometimes produces deformations with opposing directions which respects the local-discontinuities expected in movements, whereas the \textit{Jacobian Only} model typically seems to avoid penalties of the more restrictive loss term by maintaining a consistent flow direction.

Figure~\ref{fig:sample_abdomenctct_registration} (with additional examples in \ref{app:detailed_additional}) provides a broader view of the results. While both models yield similar final images, they achieve these through distinct deformation fields. A closer inspection of these fields reveals how our proposed method better preserves local discontinuities in displacements and rigidity.

\section{Conclusion}

In this paper, we have proposed novel biomechanics-inspired losses for medical image registration based on neural-networks. We have also shown that the simultaneous us of these losses to favor mechanical properties of different nature (pseudo-elasticity, rigidity and sliding constraints) at different voxel locations did not necessarily increase the complexity of the hyper-parameters tuning process, compared with the use of a homogeneous regularization loss.
A key question we asked in our introduction was also whether DLIR algorithms would be able to generalize on test data the bio-mechanically realistic deformation properties favored by our losses on training data.
Our results obtained on synthetic images and 3D medical images suggest that this is the case, which is in our opinion the major contribution of our paper.

We finally want to convey a thought related to the diffeomorphic property of the deformations, which is often desired in medical image registration.
As already show in  \cite{mok_large_2020}, integrating velocity fields that are generated by neural networks allows to preserve the topology of the structures deformed by DLIR models. Favoring spatially heterogeneous deformation properties is complex and computationally demanding using classic diffeomorphic DIR strategies. These strategies indeed ideally require to transport and to adapt the deformation properties at all discretized times of the integrated velocity field, in order to follow the deformed anatomical structures in time. The possibility to use of biomechanics-inspired losses in the fixed image space only, offered by the neural-networks training process, appeared to us as an opportunity to simply constraint diffeomorphic deformations, with locally defined mechanical properties of different nature.  

Future work will then explore the use of other bio-mechanics-inspired losses for clinically motivated applications.

\bibliographystyle{elsarticle-num} 
\onecolumn
\bibliography{references}

\appendix

\section{Strain tensor and Jacobian estimation in millimeter coordinates}
\label{app:mm_cpt}

We used $I_F \in \Omega \subset \mathbb{R}^3$ and $I_M$ to denote the fixed and moving images in the main manuscript, and considered voxel coordinates there for readability purposes.
As the voxel resolution is of primary importance when modeling physical constraints, we explain hereafter how to account for millimeter coordinates in our method.

\paragraph{Registered images alignment in millimeter coordinates}

Since $I_F$ and $I_M$ are not necessarily in the same image domain in medical imaging, it is common to model a mapping between their structures using the composition of three mappings: $I2W \circ \Phi \circ W2I$. In this general case, $W2I$ first converts millimeters coordinates of $\Omega$ into voxel coordinates, then $\Phi$ encodes the non-rigid deformations to map the structures of $I_F$ onto those of $I_M$, and then $I2W$ converts the voxel coordinates of the mapping into millimeter coordinates, plus very often allows one to rigidly align the compared images. 

We focused in the main manuscript on the estimation of $\Phi$. This means that we supposed that $I_M$ is resampled in the image domain $\Omega$ through $I2W^{-1}$. In this case converting the voxel coordinates into millimeter coordinates, is simply made by multiplying the voxel coordinates by the voxel resolutions $\delta_1$, $\delta_2$ and $\delta_3$ on the first, second and third axis, respectively.


\paragraph{Jacobian matrix and strain tensor in millimeter coordinates}

We now explain how compute the Jacobian matrix and strain tensor in millimeter coordinates, \textit{i.e.} by taking into account the scaling factors $\delta_1$, $\delta_2$ and $\delta_3$. This will be of primary importance if the voxel resolution is anisotropic. 
The equation corresponding to Eq.~\eqref{eq:jacobian_vox} is 
\begin{equation}\label{eq:jacobian_mm}
J_{mm}(x)= 
\begin{pmatrix}
                  \delta_1+     \frac{ \partial \Phi_1 (x)}{\partial x_1}& \frac{\delta_1}{\delta_2} \frac{\partial \Phi_1 (x)}{\partial x_2} & \frac{\delta_1}{\delta_3}\frac{\partial \Phi_1 (x)}{\partial x_3} \\
\null&&\\                                       
\frac{\delta_2}{\delta_1} \frac{\partial \Phi_2 (x)}{\partial x_1} &              \delta_2+         \frac{\partial \Phi_2 (x)}{\partial x_2} & \frac{\delta_2}{\delta_3}\frac{\partial \Phi_2 (x)}{\partial x_3}\\
&&\\
\frac{\delta_3}{\delta_1} \frac{\partial \Phi_3 (x)}{\partial x_1} & \frac{\delta_3}{\delta_2}\frac{\partial \Phi_3 (x)}{\partial x_2} &        \delta_3+    \frac{\partial \Phi_3 (x)}{\partial x_3}
\end{pmatrix}  \,,
\end{equation}
and the equation corresponding  to Eq.~\eqref{eq:straintensors} is:
\begin{equation}\label{eq:strain_tensors_mm}
S_{mm}(x)= 
\begin{pmatrix}
\frac{\partial u_1 (x)}{\partial x_1}   & \frac{1}{2} \left( \frac{\partial \delta_1 u_1 (x)}{\partial \delta_2 x_2} + \frac{\partial \delta_2 u_2 (x)}{\partial \delta_1 x_1} \right) & \frac{1}{2} \left( \frac{\partial \delta_1 u_1 (x)}{\partial \delta_3 x_3} + \frac{\partial \delta_3 u_3 (x)}{\partial \delta_1 x_1} \right) \\
\frac{1}{2} \left( \frac{\partial\delta_2 u_2 (x)}{\partial \delta_1 x_1} + \frac{\partial\delta_1 u_1 (x)}{\partial \delta_2 x_2} \right)  & \frac{\partial u_2 (x)}{\partial x_2} & \frac{1}{2} \left( \frac{\partial\delta_2 u_2 (x)}{\partial \delta_3 x_3} + \frac{\partial \delta_3 u_3 (x)}{\partial \delta_2 x_2} \right)\\
\frac{1}{2} \left( \frac{\partial \delta_3 u_3 (x)}{\partial \delta_1 x_1} + \frac{\partial\delta_1 u_1 (x)}{\partial \delta_3 x_3} \right)   & \frac{1}{2} \left( \frac{\partial \delta_3 u_3 (x)}{\partial \delta_2 x_2} + \frac{\partial\delta_2 u_2 (x)}{\partial \delta_3 x_3} \right) & \frac{\partial  u_3 (x)}{\partial x_3} 
\end{pmatrix}  \,.
\end{equation}

The rest of the paper is unchanged in millimeter coordinates.

\section{Viability of the loss Eq.~\eqref{eq:rigidloss} for the rotation of rigid structures}
\label{app:rigidloss}

Most image registration methods, which model locally rigid deformations explicitly consider the deformation centers $c$. We show hereafter that our loss Eq.~\eqref{eq:rigidloss}, which does not accounts for $c$, is pertinent to constrain the voxels into an image structure with rigid deformations.   To to so, we show that the local deformation variations of all points $x$ in a structure rotating around $c$ are the same as if they were rotatating around themselves. We model the rotation of $x=(x_1,x_2,x_3)$ around $c=(c_1,c_2,c_3)$ as:
\begin{equation}
\begin{pmatrix}
x_1'\\
x_2'\\
x_3'\\
1
\end{pmatrix}
=
\begin{pmatrix}
1&0&0&c_1'\\
0&1&0&c_2'\\
0&0&1&c_3'\\
0&0&0&1
\end{pmatrix}
\left(
\begin{array}{ccc|c}
&&&0\\
&R&&0\\
&&&0\\ \hline
0 & 0 & 0 & 1 \\
\end{array}
\right)
\begin{pmatrix}
1&0&0&-c_1'\\
0&1&0&-c_2'\\
0&0&1&-c_3'\\
0&0&0&1
\end{pmatrix} \,,
\end{equation}
where $R$ is a 3D rotation matrix. Developing this computation shows that
\begin{equation}
\begin{pmatrix}
x_1'\\
x_2'\\
x_3'
\end{pmatrix}
=
\begin{pmatrix}
\null&\null&\null\\
\null&R&\null\\
\null&\null&\null\\
\end{pmatrix} 
\begin{pmatrix}
x_1\\
x_2\\
x_3
\end{pmatrix} 
-
\begin{pmatrix}
\null&\null&\null\\
\null&R&\null\\
\null&\null&\null\\
\end{pmatrix} 
\begin{pmatrix}
c_1\\
c_2\\
c_3
\end{pmatrix} 
+
\begin{pmatrix}
c_1\\
c_2\\
c_3
\end{pmatrix} \,.
\end{equation}
As a consequence,  $x$ is rotated with $R$ and translated with $c-R c^{\intercal}$. 
When represented by a strain tensor, which does not account for translations, its local deformation variations are then the same as if the rotation center was local. The value of the  loss Eq.~\eqref{eq:rigidloss} is then zero.

\section{Definition of the loss weights}\label{sec:loss_weights}

In practice, we use  $loss_{rigid}$ and $loss_{shearing}$ at pertinent voxel locations only and $loss_{Jac}$ otherwise. This means that losses of different nature are used simultaneously at different voxel locations. An important practical question is then how to weight these different losses to ensure that they have a similar impact on the deformation smoothness level.
We actually designed these losses, so that they smooth the deformations at similar scales if they have similar weights. 
%
We then discuss in this subsection why $loss_{Jac}$ should be tuned as similar scales as $loss_{rigid}$ and $loss_{shearing}$ for similar levels of deformation variations. 

We first remark that $loss_{shearing}$ is a truncated version of $loss_{rigid}$ with respect to a linear projection of $u$. We then focus on a comparison between $loss_{rigid}$ and $loss_{Jac}$.
Let us  consider the eigenvalues $(\lambda_1^J,\lambda_2^J,\lambda_3^J)$ of $J(x)$. It can be remarked that these $\lambda_i^J$ are equal to $1$ if there is no deformation variation around $x$, while the eigenvalues $\lambda_i$ of the strain tensor in Eq.~\eqref{eq:eigen_decomp} are equal to $0$ in the same case. It  can also be easily shown that $det(J(x))= \prod_{i=1}^3 \lambda_i^J$, and then $\log{(det(J(x)))} = \sum_{i=1}^3  \log{(\lambda_i^J)}$. Using first order Taylor expansions around the  $\lambda_i^J = 1$, representing locally stationary deformations, we can also write $\log{(det(J(x)))} \approx \left. \sum_{i=1}^3  (\lambda_i^J - 1) \right|_{\lambda_i^J \approx 1 \,, i=1,2,3}$. 
For reasonably smooth variations of $u$, and therefore  $\lambda_i^J \approx 1$ and $\lambda_i \approx 0$,  the variations of $\log{(det(J(x)))}$ should then have a similar amplitude as those of $ \sum_{i=1}^3  \lambda_i$. Both $loss_{rigid}$ and $loss_{Jac}$ are also quadratic loss with respect to these terms and should finally vary at similar scales. 

It therefore appears as reasonable to weight the losses $loss_{rigid}$, $loss_{shearing}$, and $loss_{Jac}$ with the same weights when reasonably smooth deformations are expected. We at least know that the optimal weights to penalize similarly the deformations in areas with supposed rigid, smooth or sliding motion should have similar values. This is particularly helpful when defining the registration hyper-parameters.

\section{Regularization Regions Overview}
\label{app:reg_regions}

Through discussions with radiologists and the use of anatomy textbooks, we compiled two lists of organs whose segmentation masks were used to defined different types of expected deformations, hence different losses as was described in Section ~\ref{sec:annot_images}. 

A rigidity loss was applied to bone tissues using the following labels:
\textbf{Vertebrae} (S1, L5, L4, L3, L2, L1, T12, T11, T10, T9, T8, T7, T6, T5, T4, T3, T2, T1, C7, C6, C5, C4, C3, C2, C1) ; \textbf{Ribs} (left and right : 1, 2, 3, 4, 5, 6, 7, 8, 9, 10, 11, 12) ; \textbf{Humerus} (left and right) ; \textbf{Scapula} (left and right) ; \textbf{Clavicula} (left and right) ; \textbf{Femurs} (left and right) ; \textbf{Hips} (left and right) ; \textbf{Sacrum} ; \textbf{Skull}.

\begin{figure}[ht]
    \centering
    \includegraphics[width=0.7\linewidth]{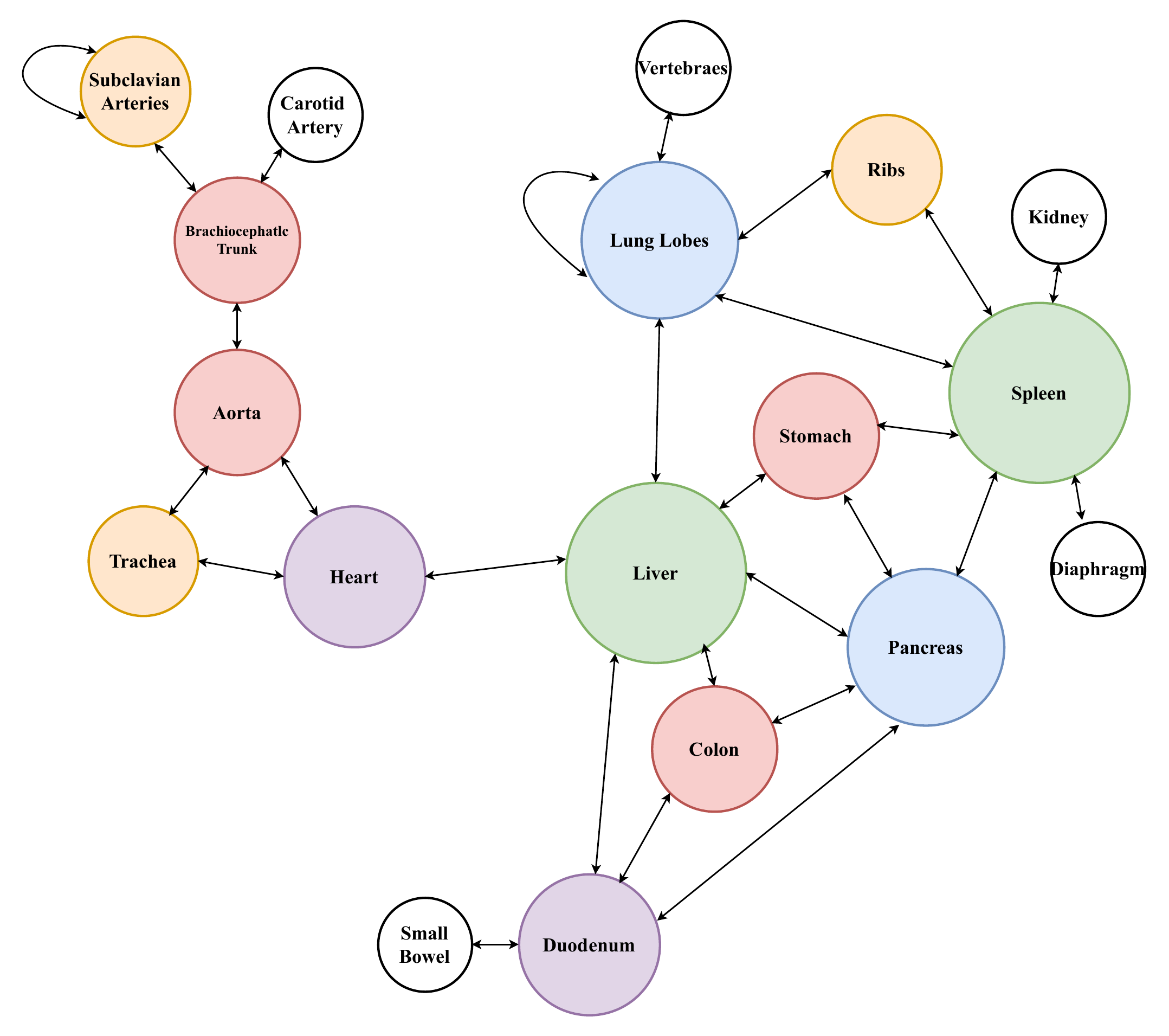}
    \caption{Graph of tissue interfaces between which Strain is permitted}
    \label{fig:appendix_labels_strain}
\end{figure}

Meanwhile, through the process detailed in Section ~\ref{ssec:reg_mask}, a strain prone region was allowed on the interfaces of the tissues pairs identified by radiologists as described in Fig.~\ref{fig:appendix_labels_strain}. Note that for readability, different sub-structures of tissues where confounded into single nodes on the graph (lung lobes, vertebrae, kidneys, ribs, etc...). Also note that nodes with self-edges (e.g "Lung Lobes") are due to interactions between sub-structures of the tissue (such as lung lower and upper lobes due to pleura contact).

\section{Additional results obtained on 3D images of the AbdomenCTCT dataset}
\label{app:detailed_additional}

\begin{figure}[htb]
         \centering
        \includegraphics[width=1.0\linewidth]{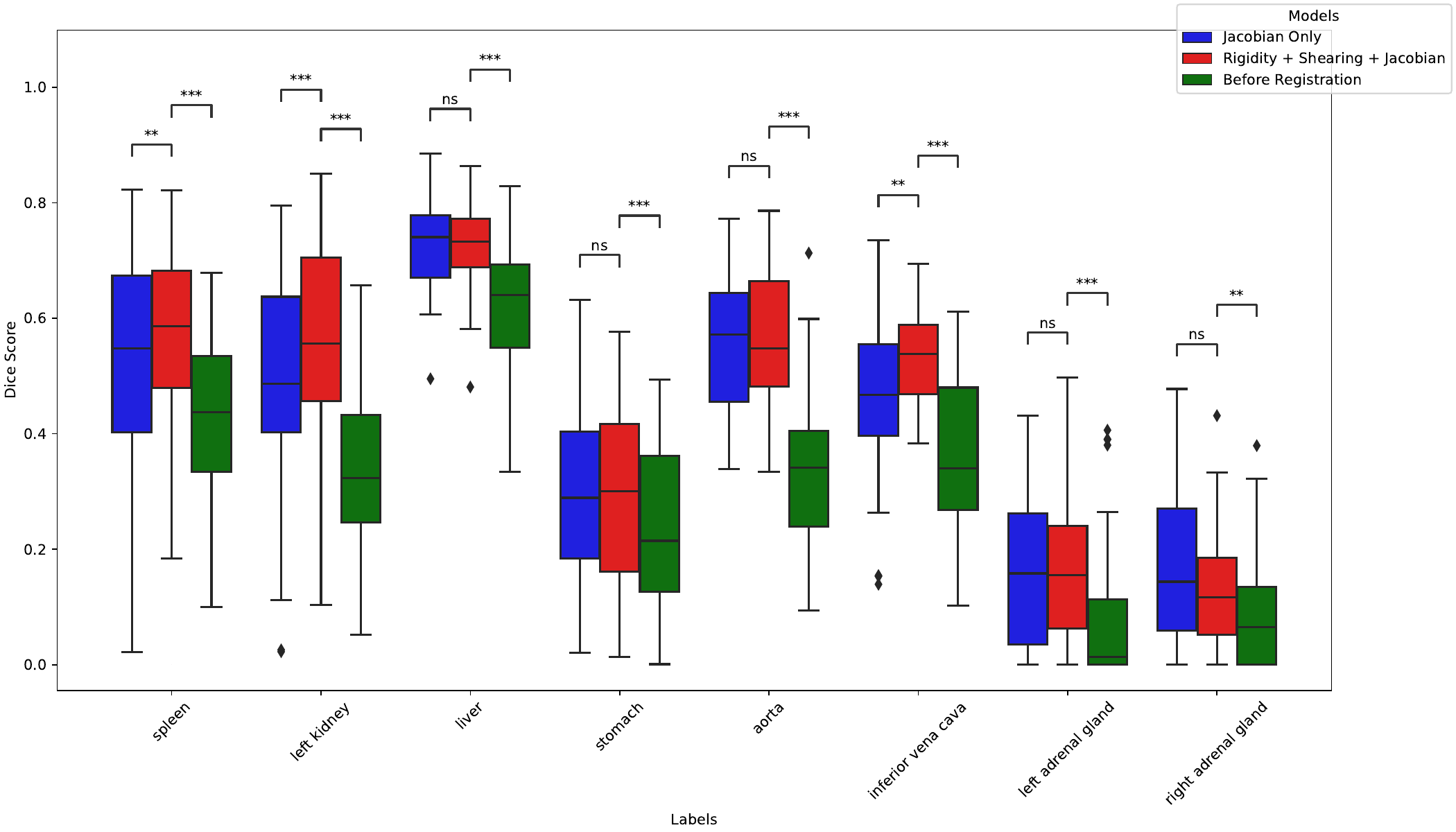}

        \caption{Detailed Dice scores per organ. Differences in average metrics are compared by a Wilcoxon signed-rank test with the following tiers: $n.s \to \textit{Not Significant}, \ast \to p<10^{-1},\ast\ast \to p<10^{-2} , \ast\ast\ast \to p<10^{-3} $ }
        \label{fig:detailed_dice}
\end{figure}

\begin{figure}[htb]
         \centering
        \includegraphics[width=0.\linewidth]{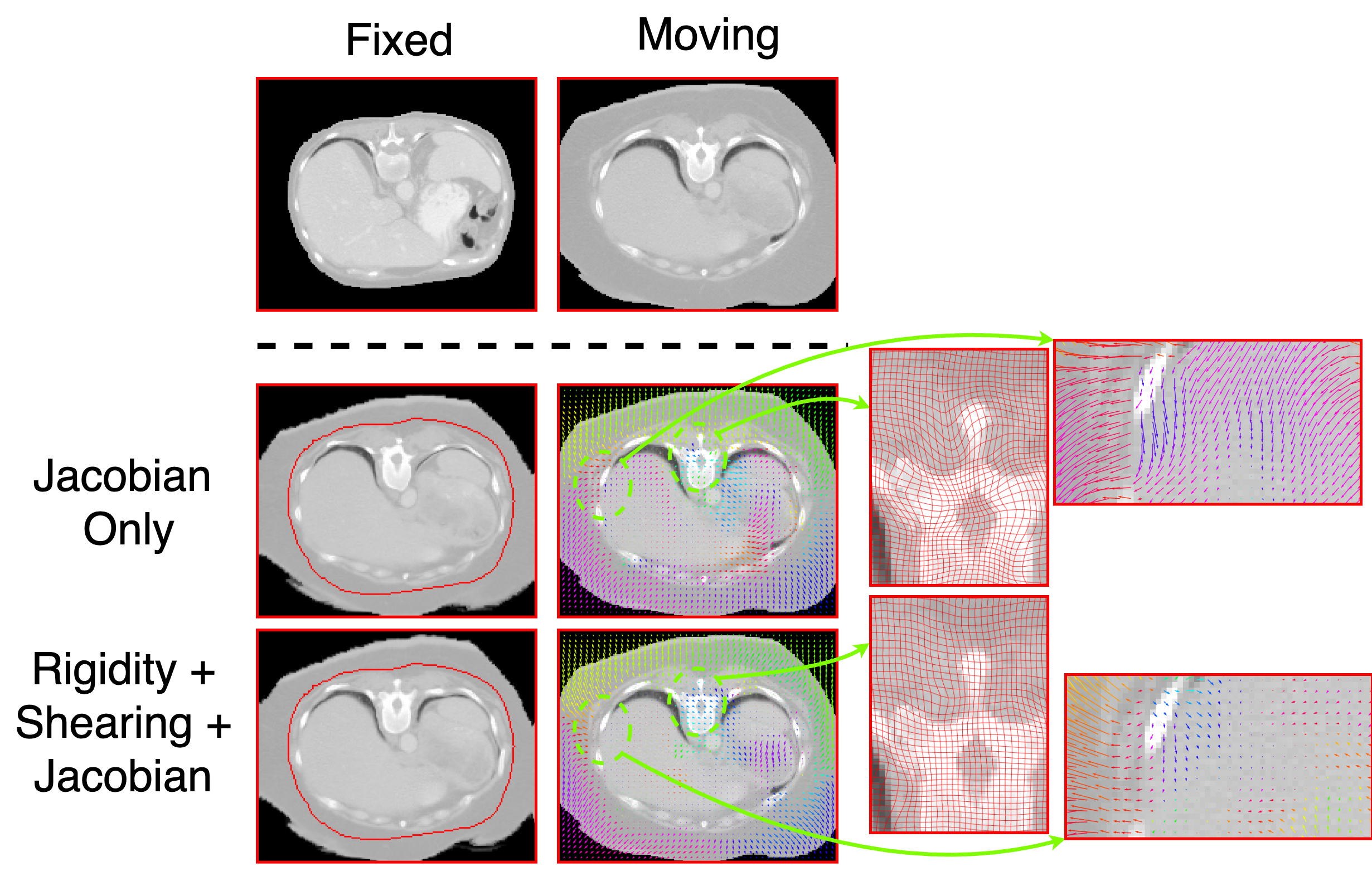}

        \caption{Complementary illustration to Fig.~\ref{fig:sample_abdomenctct_registration} obtained on patient 2}
        \label{fig:sample_abdomenctct_registration2}
\end{figure}

\begin{figure}[htb]
         \centering
        \includegraphics[width=0.7\linewidth]{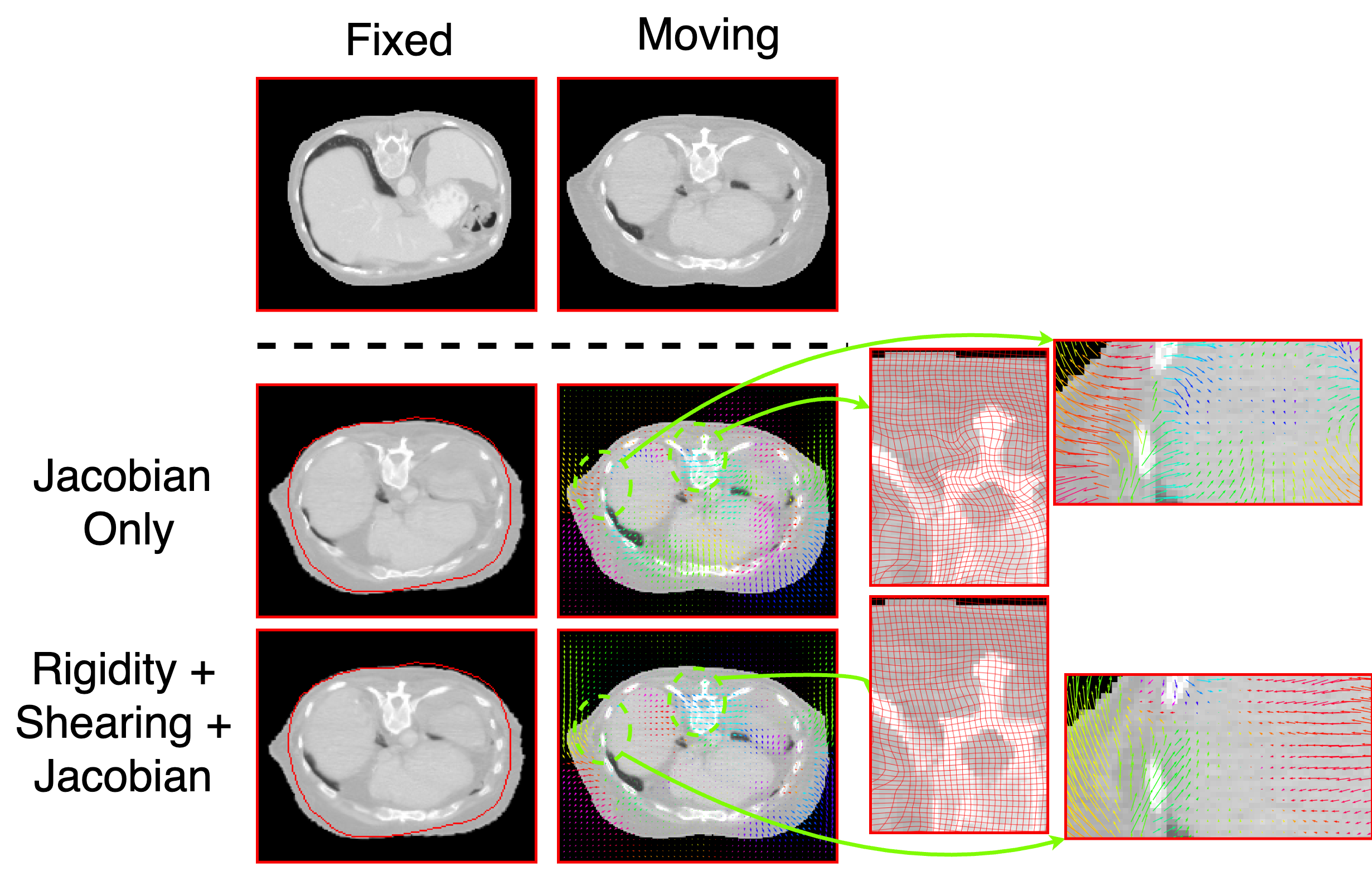}

        \caption{Complementary illustration to Fig.~\ref{fig:sample_abdomenctct_registration} obtained on patient 3}
        \label{fig:sample_abdomenctct_registration3}
\end{figure}

\begin{figure}[htb]
         \centering
        \includegraphics[width=0.7\linewidth]{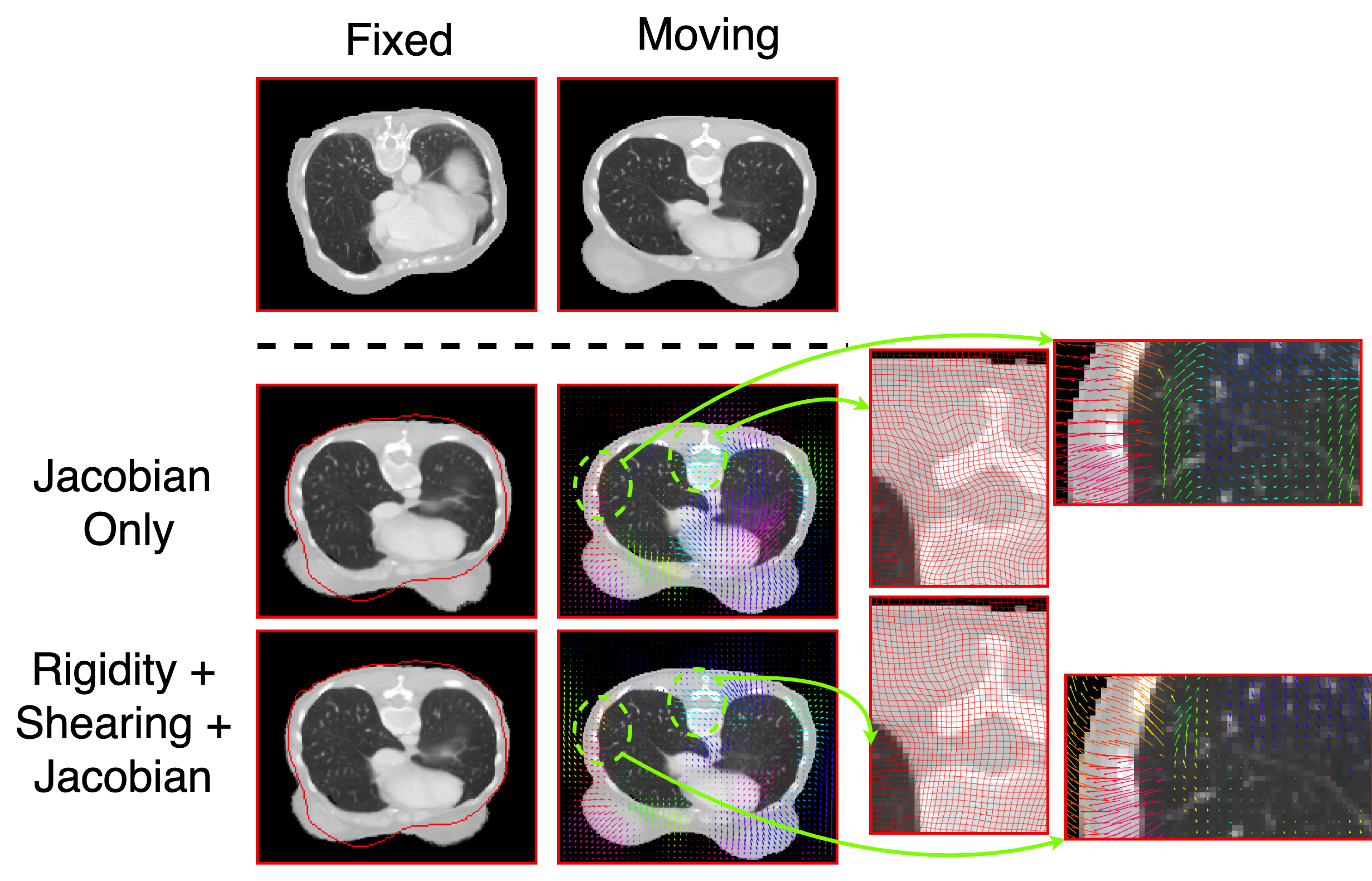}

        \caption{Complementary illustration to Fig.~\ref{fig:sample_abdomenctct_registration} obtained on patient 4}
        \label{fig:sample_abdomenctct_registration4}
\end{figure}

We give in this last appendix different results we obtained that complement the results of Subsection~\ref{ssec:ResRealData}.
The Dice scores for the different annotations provided in the official Learn2Reg AbdomenCTCT dataset are summurized Figure~\ref{fig:detailed_dice}.
Additional sample registrations on the AbdomenCTCT challenge can also be found in Figures~\ref{fig:sample_abdomenctct_registration2}, \ref{fig:sample_abdomenctct_registration3} and \ref{fig:sample_abdomenctct_registration4}.

\end{document}